\documentclass[preprint]{elsarticle}

\usepackage{charter}
\usepackage{amsmath,amsfonts,amssymb,amsthm}
\usepackage[euler-digits,euler-hat-accent]{eulervm}
\usepackage{bm}
\usepackage{lineno,hyperref}
\usepackage{caption}
\usepackage{subcaption}
\usepackage{float}
\usepackage{tikz}
\usetikzlibrary{shapes.geometric, arrows}
\usepackage{empheq}
\modulolinenumbers[1]
\usepackage{enumitem}
\usepackage{graphicx}
\usepackage{array}
\usepackage{stackrel}
\usepackage{relsize}
\newcommand{\F}[1]{\mathbb{#1}}
\newcommand{\C}[1]{\mathcal{#1}}
\def\Jac{{{J(\bm{q})}}}
\newcommand*{\horzbar}{\rule[.5ex]{3.5ex}{0.5pt}}
\newcommand{\commentedbox}[2]{%
	\mbox{
		\begin{tabular}[t]{@{}c@{}}
			$\boxed{\displaystyle#1}$\\[1mm]
			$#2$
		\end{tabular}%
	}%
}
\newtheorem{theo}{Theorem}[section]
\newtheorem{cor}[theo]{Corollary}
\newtheorem{lem}[theo]{Lemma}
\newtheorem{pro}[theo]{Proposition}
\theoremstyle{definition}
\newtheorem{defi}[theo]{Definition}
\theoremstyle{remark}
\newtheorem{rem}[theo]{Remark}
\newtheorem*{ejem}{Example}

\makeatletter
\def\ps@pprintTitle{%
	\let\@oddhead\@empty
	\let\@evenhead\@empty
	\def\@oddfoot{}%
	\let\@evenfoot\@oddfoot}
\makeatother

\begin{document}
	
\begin{frontmatter}
		
		\title{Singularities of serial robots: Identification and distance computation using geometric algebra}
		
		
		\author[mymainaddress]{Isiah Zaplana\corref{mycorrespondingauthor}}
		\cortext[mycorrespondingauthor]{Corresponding author}
		\ead{isiah.zaplana@kuleuven.be}
		
		\author[mysecondaryaddress]{Hugo Hadfield}
		\ead{hh409@cam.ac.uk}
		
		\author[mysecondaryaddress]{Joan Lasenby}
		\ead{jl221@cam.ac.uk}
		
		\address[mymainaddress]{Department of Mechanical Engineering, KU Leuven, Leuven, Belgium}
		\address[mysecondaryaddress]{Department of Engineering, University of Cambridge, Cambridge, UK}
		
		\begin{abstract}
			The singularities of serial robotic manipulators are those configurations in which the robot loses the ability to move in at least one direction. Hence, their identification is fundamental to enhance the performance of current control and motion planning strategies. While classical approaches entail the computation of the determinant of either a $6\times n$ or $n\times n$ matrix for an $n$ degrees of freedom serial robot, this work addresses a novel singularity identification method based on modelling the twists defined by the joint axes of the robot as vectors of the six-dimensional and three-dimensional geometric algebras. In particular, it consists of identifying which configurations cause the exterior product of these twists to vanish. In addition, since rotors represent rotations in geometric algebra, once these singularities have been identified, a distance function is defined in the configuration space $\mathcal{C}$ such that its restriction to the set of singular configurations $\mathcal{S}$ allows us to compute the distance of any configuration to a given singularity. This distance function is used to enhance how the singularities are handled in three different scenarios, namely motion planning, motion control and bilateral teleoperation.
		\end{abstract}
		
		
\end{frontmatter}

\section{Introduction}\label{Intro}
A \textit{serial robot manipulator} is an open kinematic chain made up of a sequence of rigid bodies, called \textit{links}, connected by means of actuated kinematic pairs, called \textit{joints}, that provide relative motion between consecutive links. At the end of the last link, there is a tool or device known as the \textit{end-effector}. Only two types of joints are considered throughout this work: \emph{revolute joints}, that only perform rotations, and \emph{prismatic joints}, that only perform translations. If joint $i$ is revolute (prismatic), the amount it rotates (translates) is encoded by an angle $\theta_i$ (a displacement $d_i$). These scalars are known as \textit{joint variables} of the robot.

From a kinematic point of view, the end-effector position and orientation (also known as the \emph{pose}) can be expressed as a differentiable function $f:\mathcal{C}\to X$, where $\mathcal{C}$ denotes the space of joint variables, called the \textit{configuration space} of the robot, and $X$, the space of all positions and orientations of the end-effector with respect to a reference frame, which is usually called the \textit{operational space}. A serial robot is said to have \textit{$n$ degrees of freedom} (DoF) if its configuration can be minimally specified by $n$ variables. For a serial robot, the number and nature of the joints determine the number of DoF. For the task of positioning and orientating its end-effector in the three-dimensional space, the manipulators with more than 6 DoF are called \emph{redundant} while the rest are \emph{non-redundant}.

\begin{figure}[t!]
	\centering
	\includegraphics[width=6cm]{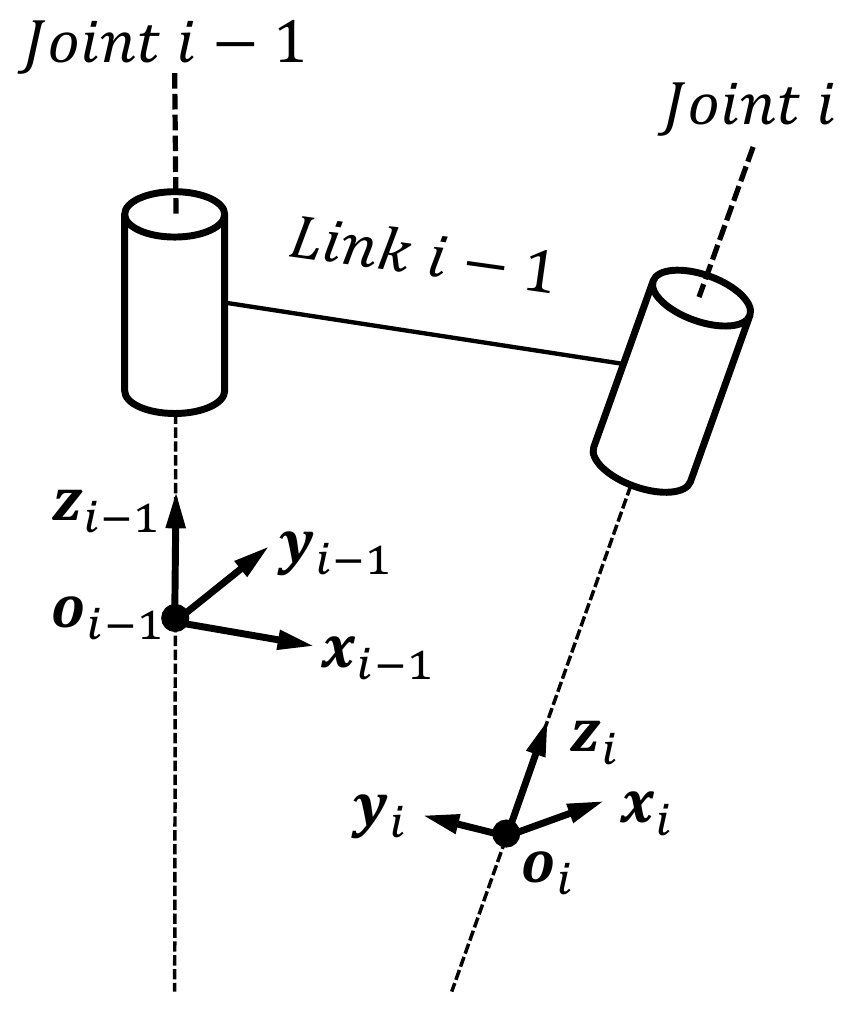}
	\caption{A frame $\{\bm{o},\bm{x},\bm{y},\bm{z}\}$ is attached to each joint of the serial robot to describe its relative position and orientation.}
	\label{DH-SJ}
\end{figure}

In order to describe its relative position and orientation, a frame $\{\bm{o},\bm{x},\bm{y},\bm{z}\}$ is attached to each joint (figure \ref{DH-SJ}). The relations between consecutive joint frames are conventionally described by homogeneous transformation matrices. In particular, $^{i-1}T_{i}$ relates frame $\{i\}$ to frame $\{i-1\}$ (the first joint frame is related to a fixed reference frame, known as the \textit{world frame}). Therefore, the end-effector pose $^{0}T_{n}$ of a robot with $n$ DoF with respect to the world frame can be represented as:
\begin{equation}\label{RelationMatrices}
	^{0}T_{n} = {}^{0}T_{1}\: {}^{1}T_{2}\;\cdots\; {}^{n-1}T_{n}
\end{equation}
with
\begin{equation}\label{frepresentation}
	^{0}T_{n} = \begin{pmatrix}
		R & \bm{p}\\
		0 & 1
	\end{pmatrix},
\end{equation}
where $R$ is a rotation matrix that describes the end-effector orientation with respect to the world frame, while $\bm{p}$ is a position vector describing the end-effector position with respect to the world frame. This description is equivalent to the one provided by $f$, known as the \textit{kinematic function} of the serial robot. Thus, $f(\bm{q}) = \bm{x}$, where $\bm{x}$ denotes the vector describing the end-effector pose and $\bm{q} = (q_{1},\dots,q_{n})$ denotes the vector whose components are the joint variables, also known as the \emph{configuration} of the robot. Clearly, either $q_{i} = \theta_{i}$ if joint $i$ is revolute or $q_{i} = d_{i}$ if joint $i$ is prismatic. 

Deriving the kinematic relation defined by $f$ with respect to time, we obtain another relation:
\begin{equation}\label{kinematicrelation}
	\dot{\bm{x}} = J(\bm{q})\dot{\bm{q}},
\end{equation}
where $\dot{\bm{x}}$ denotes the end-effector velocity vector; $\dot{\bm{q}}$, the vector of the joint velocities and $J$, the \textit{Jacobian matrix} of $f$. If $J = [J_{1}\,\cdots\,J_{n}]$, then each column $J_i$ can also be computed as:
\begin{equation}\label{Jacobian}
	\begin{array}{ll}
		J_{i} = \left[\begin{array}{c}
			\bm{z}_{i}\times(\bm{o}_{n}-\bm{o}_{i})\\
			\bm{z}_{i}
		\end{array}\right] &\quad \text{if joint }i\text{ is revolute},\\[2ex]
		J_{i} = \left[\begin{array}{c}
			\bm{z}_{i}\\
			0
		\end{array}\right] &\quad \text{if joint }i\text{ is prismatic}.
	\end{array}
\end{equation}

\begin{defi}\label{sing_def}
	Given a serial robot with $n$ DoF, a \textit{singularity} or \textit{kinematic singularity} is a configuration $\bm{q}\in\mathcal{C}$ satisfying $\rho(J(\bm{q}))<\min\{n,6\}$, where $\rho(\cdot)$ denotes the rank of the matrix argument. The set of all singular configurations is a subset of $\mathcal{C}$ that is usually denoted by $\mathcal{S}$ and known as the \textit{singular set}.
\end{defi}
Using the relation (\ref{kinematicrelation}), it is easy to see that if $\bm{q}\in C$ is a singularity of a given serial robot, then the following two statements hold:
\begin{itemize}
	\item The robot loses at least one degree of freedom or, equivalently, its end-effector cannot be translated or rotated around at least one Cartesian direction.
	\item Finite linear and angular velocities of the end-effector may require infinite joint velocities.
\end{itemize}
In addition, Gottlieb \cite{Gottlieb86} and Hollerbach \cite{Hollerbach85} have independently proven that any serial manipulator with $n>2$ DoF has singularities. The identification of such singularities is made by solving the following non-linear equation:
\begin{equation}\label{determinantJacobian}
	\det(J(\bm{q})) = 0
\end{equation}
if the robot is non-redundant, and
\begin{equation}\label{determinantJacobianredundant}
	\det(J(\bm{q})J^{T}(\bm{q})) = 0
\end{equation}
if it is redundant.

In general, if the serial robot possesses at least one revolute joint, several coefficients of the Jacobian matrix are non-linear expressions and, thus, neither equation \eqref{determinantJacobian} nor equation \eqref{determinantJacobianredundant} are easy to formulate and solve. However, for manipulators with a spherical wrist, a simplification can be made. For these robots, the axes of their last three joints intersect at a common point, known
as the \textit{wrist center point}, or are parallel (the intersection point and, hence, the wrist center point, is the point at the infinity). Since the origin of the frame attached to the end-effector can be placed at the wrist center point, a zero block appears in $J(\bm{q})$ by definition (see equation \eqref{Jacobian}). Hence:
\begin{equation}\label{Bloques}
	J(\bm{q}) = \begin{bmatrix}
		J_{11}(\bm{q}) & 0\\
		J_{21}(\bm{q}) & J_{22}(\bm{q})
	\end{bmatrix},
\end{equation}
where $J_{11}(\bm{q}),J_{21}(\bm{q})$ are blocks of order $3\times(n-3)$ and $J_{22}(\bm{q})$ is a block of order $3$. Now, equation \eqref{determinantJacobian} is simplified to:
\begin{equation}\label{determinantdecoupled}
	\det(J(\bm{q})) = \det(J_{11}(\bm{q}))\det(J_{22}(\bm{q})),
\end{equation}
from which the singularities can be obtained as solutions of either $\det(J_{11}(\bm{q})) = 0$ or $\det(J_{22}(\bm{q})) = 0$. These two equations allow us to decouple the singularities into position and orientation singularities as follows:
\begin{itemize}
	\item Position singularities $PS = \{\bm{q}\in C\,:\,\det(J_{11}(\bm{q})) = 0\}$.
	\item Orientation singularities $OS = \{\bm{q}\in C\,:\,\det(J_{22}(\bm{q})) = 0\}$.
\end{itemize}
Similarly, we can make the same decoupling for redundant robots:
\begin{itemize}
	\item Position singularities $PS = \{\bm{q}\in C\,:\,\text{rank}(J_{11}(\bm{q})) <3\}$.
	\item Orientation singularities $OS = \{\bm{q}\in C\,:\,\det(J_{22}(\bm{q})) = 0\}$.
\end{itemize}

\begin{rem}
	The Jacobian matrix $\Jac$ is represented with respect to the world frame (usually located at the base of the robot). However, sometimes it is useful to represent $\Jac$ in a different frame $\mathcal{B}$. To do so, the following identity is used:
	\begin{equation}\label{B}
		\Jac^{\mathcal{B}} = B\Jac,
	\end{equation}
	where:
	\begin{equation}
		B = \begin{bmatrix}
			R_{0}^{\mathcal{B}} & 0\\
			0 & R_{0}^{\mathcal{B}}
		\end{bmatrix},
	\end{equation}
	with $R_{0}^{\mathcal{B}} = (R_{\mathcal{B}}^{0})^{T}$ and where $R_{\mathcal{B}}^{0}$ denotes the rotation matrix that relates the orientation of $\mathcal{B}$ with respect to the orientation of the world frame. 
\end{rem}
Singularity identification is one of the fundamental research fields in robot kinematics since, as stated before, they affect the motion of the robot and its performance when executing different tasks. Therefore, such identification is fundamental to enhancing the performance of current control and motion planning strategies by designing approaches to handle them. For instance, current applications of the subject include the handling of singularities for a robust control architecture in human-robot collaboration \cite{Carmicheletal2020}, the planning of singularity-free trajectories in robot-assisted surgery \cite{Thananjeyanetal2019} and the smooth trajectory generation for rotating extensible manipulators or painting robots \cite{Dupac2018,Wangetal2018}. However, such identification is still problematic. The majority of the current approaches are based on the computation of either $\det(\Jac) =0$ or $\det(\Jac J^{T}(\bm{q}))=0$ or on the manipulation of the singular values of $\Jac$ \cite{Ratajczak2020,Almarkhi2019,sharifi2019identification} and, hence, they are not computationally efficient. In addition, there is no efficient way of computing how close an arbitrary configuration is to a given singularity (which is fundamental for defining a threshold from where the strategies to handle them start to work). In this context, geometric algebra turns out to be very useful. In addition, it is currently applied to several problems in robot kinematics and geometry \cite{Hugo2020,Thiruvengadam2021}.

There is not much literature regarding the identification of singularities using geometric algebra and the majority of the contributions focus on parallel mechanisms. For serial robots, Corrochano \& Sobczyk \cite{CorrochanoSobczyk01} extend the Lie bracket of two vectors defined in any Lie algebra to what they call the \textit{superbracket} of the lines $\ell_{1},\dots,\ell_{6}$, $[\ell_{1},\dots,\ell_{6}]$, where the line $\ell_{i}$ denotes the axis of joint $i$. For serial robots with 6 DoF, the main idea is to split the superbracket into small superbrackets, called bracket monomials, that are equated to zero. The singularities are the solutions of these bracket monomial equations. Following the same idea, Kanaan et al. \cite{KanaanWengerCaroChablat09} define the superbracket in a Grassmann-Caley algebra. Since the Lie bracket is well defined in every Grassmann-Caley algebra, the superbracket is also well defined. However, splitting the superbracket into the bracket monomials in this context is not intuitive and, as a consequence, is not always realizable. In addition, there is not standard procedure for computing these brackets monomials.

For parallel mechanisms, the majority of works \cite{Tanev06,ChaiXiang16,YaoChenChaiLi17,ChaiLi17,Maetal2017} focus on approaches developed for some particular parallel robots. The main idea consists of computing, for each leg of the mechanism, the exterior product of the twists defined by its joints and equating it to zero. For those legs with less than six actuated joints, combinations of two, three or more legs are considered. The main problem with these approaches is their lack of generalisability. Each approach is designed for the specific parallel robot the authors work with.

Huo et al. \cite{HuoSunSong17} present a mobility analysis applying conformal geometric algebra, and a singularity analysis using an idea similar to the ones presented in the above-mentioned contributions. A mobility analysis of overconstrained parallel mechanisms is performed using Grassmann-Cayley algebra by Chai et al. \cite{Chai2017}, while Yang and Li \cite{Yang2020} propose a novel identification method for the constraint singularities of parallel robots based on differential manifolds. Finally, Kim et al. \cite{KimJeongPark15a} apply conformal geometric algebra to the identification of the singularities of a particular type of parallel manipulator, the SPS-parallel manipulator. Several lines and planes are defined using the different joint axes. Then, the relative positions of different combinations of these geometric entities are studied to geometrically find the singularities. However, this method cannot be extended to other classes of parallel or serial robots nor can it be implemented as an algorithm due to its complex geometrical nature. 

In this paper, a novel approach for singularity identification based on the six-dimensional and three-dimensional geometric algebras is introduced. It extends the works developed for parallel robots and reviewed above. In particular, one of the novelties of this method is that it can be applied to both redundant and non-redundant serial robots of any geometry. We first model the twists defined by the joint axes as vectors of the six-dimensional geometric algebra and, then, we manipulate the exterior product of these twists. In addition, this method can be simplified for serial robots with a spherical wrist using, instead of the six-dimensional geometric algebra $\mathcal{G}_6$, the three-dimensional geometric algebra $\mathcal{G}_3$. Once the singularities have been identified and since rotors describe the transformations between arbitrary multivectors in geometric algebra, a distance function $D$ can be defined in the configuration space $\mathcal{C}$ that can be used to determine the distance of any arbitrary configuration $\bm{q}\in\mathcal{C}$ to a given singularity $\bm{q}_{s}\in\mathcal{S}$. This is the first time, to the best of the authors' knowledge, that such a distance function has been defined. It is well-known that there are several indexes that can be used to check whether a given configuration is close or not to a singularity. For instance, the Jacobian matrix $\Jac$ allows us to define the \textit{manipulability index} $w_{\text{m}}$ as:
\begin{equation}
	w_\text{m} = \sqrt{\det(\Jac^T\Jac)} = \sigma_n\cdots\sigma_1,
\end{equation}
where $0\leq\sigma_1\leq\cdots\leq\sigma_n$ are the singular values of $\Jac$. Alternatively, we can also define the \textit{condition number} of $\Jac$, $w_\text{c} = \sigma_n/\sigma_1$. Clearly, the former is close to zero when the configuration is close to a singularity, while the value of the latter increases as the robot approaches to a singular configuration. Although there are several approaches based on the use of such indexes \cite{Huo2008,Yahya2012}, none of them defines a distance function and, as stated in \cite{Siciliano08a}, they do not provide a realistic measure of how close a singularity is, just whether it is close or not. On the other hand, Yao et al. \cite{Yao2018} propose a different index of closeness to singularities for planar parallel robots based on the volume of the workspace. Despite interesting, it is still not a distance function and it can neither be easily applied to serial robots. Similarly, Nawratil \cite{Nawratil2019} defines a distance function for parallel manipulators of the Stewart-Gough type. However, it measures how close a given pose of the end-effector is to a singular pose (i.e., the pose associated with a singular configuration). Hence, such a distance function is not defined in the configuration space $\mathcal{C}$ but in the operational space $X$. Finally, Bu \cite{Bu_2016} defines an angle between the velocity vector associated with one of the joints and the manifold generated by the others. Again, such an angle acts as a measure of closeness but not as a distance function and, thus, it does not provide a realistic measure of how close a singularity is.

The rest of the paper is organized as follows: Section \ref{Background} presents an overview of geometric algebra that will be useful for understanding the proposed contribution. In Section \ref{IdentificationSing}, the novel singularity identification approach and the simplification for serial robots with a spherical wrist are fully developed, while the novel distance function is constructed in Section \ref{Distance}. The application of these results to the Kuka LWR 4+,  a redundant serial robot with a spherical wrist, is given in Section \ref{ApplicationKuka}.  Section \ref{Handling} lists three different applications where both the singularity identification and the novel distance function can be applied in order to illustrate their utility. Finally, the conclusions are given in Section \ref{Conclusions}.

\section{Mathematical preliminaries: Geometric algebra}\label{Background}
One of the main problems of vector spaces is that linear transformations between them are represented through matrices which entails a high computational cost when implemented. To overcome these and related problems, geometric algebra provides an excellent framework. Throughout this section, a brief overview of geometric algebra is presented. More detailed treatments of the subject can be found in \cite{DoranLasenby03, Dorst-Fontijne-Mann-2007}.

\begin{defi}\label{exteriorproduct}
	Given two vectors $\bm{x}_{1},\bm{x}_{2}\in\F{R}^{n}$, the \textit{outer} or \textit{exterior product} of $\bm{x}_{1}$ and $\bm{x}_{2}$, $\bm{x}_{1}\wedge \bm{x}_{2}$, is a new element that can be seen as the oriented area of the parallelogram obtained by sweeping the vector $\bm{x}_{1}$ along $\bm{x}_{2}$. The exterior product is bilinear, associative and anticommutative. In particular, $\bm{x}\wedge\bm{x} = 0$ for every $\bm{x}\in\mathbb{R}^n$.
\end{defi}
The new element defined by the exterior product is called a \textit{bivector} and it is defined to have grade two. By extension, the outer product of a bivector with a vector is known as a \textit{trivector}, is denoted by $\bm{x}_{1}\wedge\bm{x}_{2}\wedge \bm{x}_{3}$ and defined to have grade three. Trivectors can be seen as the oriented volume obtained by sweeping the bivector $\bm{x}_{1}\wedge\bm{x}_{2}$ along $\bm{x}_{3}$.

This can be generalized to an arbitrary dimension. Thus
\begin{equation}\label{rblade}
	\bm{x}_{1}\wedge \bm{x}_{2} \wedge \dots \wedge \bm{x}_{k}
\end{equation}
denotes a \textit{$k$-blade}, i.e., an element of grade $k$. Linear combinations of $k$-blades are known as \textit{$k$-vectors}, while linear combinations of $k$-vectors (for different $k$) are known as \textit{multivectors}.

In his work \cite{Clifford82}, Clifford extends the exterior product by adding a scalar product between vectors, the \textit{inner product}. He defines the \textit{geometric product} (also known as the \textit{Clifford product}) as follows:
\begin{equation}\label{geometricproduct}
	\bm{x}_{1}\bm{x}_{2} = \bm{x}_{1}\cdot\bm{x}_{2}+\bm{x}_{1}\wedge \bm{x}_{2}\quad(\bm{x}_{1},\bm{x}_{2}\in\F{R}^{n}).
\end{equation}
Thus, the geometric product between two vectors has two components: the scalar component given by the inner product and the bivector component given by the exterior product. Clearly, it also inherits the associativity and bilinearity of the exterior product. 

When applied to an orthonormal basis $B = \{e_{1},\dots,e_{n}\}$ of $\F{R}^{n}$, the geometric product acts as follows:
\begin{equation}\label{elementsofbasis}
	e_{i}e_{j}  = \left\{\begin{array}{lcl}
		1 & \text{for} & i=j\\
		e_{i}\wedge e_{j} & \text{for} & i\neq j
	\end{array}\right.
\end{equation}
Thus, for each $0\leq k\leq n$, the set of $k$-vectors is spanned by:
\begin{itemize}[leftmargin=13mm]
	\item[\boxed{k=0}] $\{1\}$ (scalars).
	\item[\boxed{k=1}] $\{e_{1},\dots,e_{n}\}$ (vectors).
	\item[\boxed{k=2}] $\{e_{i}\wedge e_{j}\}_{1\leq i<j\leq n}$ (bivectors).
	\item[\boxed{k=3}] $\{e_{i}\wedge e_{j}\wedge e_{k}\}_{1\leq i<j<k\leq n}$ (trivectors).
	\item[\empty] $\vdots$
	\item[\boxed{k=r}] $\{e_{i_{1}}\wedge\dots\wedge e_{i_{r}}\}_{1\leq i_{1}<\dots<i_{r}\leq n}$ ($r$-vectors).
	\item[\empty] $\vdots$
	\item[\boxed{k=n}] $\{e_{1}\wedge\dots\wedge e_{n}\}$ (pseudoscalar).
\end{itemize}
Then, for each $0\leq k \leq n$, there are exactly $C(n,k)$ generators for the set of $k$-vectors and, thus, the set of $k$-vectors defines a vector space with basis $B_k = \{e_{i_{1}}\wedge\dots\wedge e_{i_{k}}\}_{1\leq i_{1}<\dots<i_{k}\leq n}$ and dimension $C(n,k)$. 

\begin{defi}
	Let $\F{R}^{n}$ denote the real vector space of dimension $n$. Then, the vector space spanned by the basis
	\begin{equation}\label{geometricalgebrabasis}
		\mathcal{B} = \{e_{i_{1}}\wedge\dots\wedge e_{i_{r}}\}_{\substack{1\leq i_{1}<\dots<i_{r}\leq n\\0\leq r\leq n}}
	\end{equation}
	endowed with the geometric product defined in (\ref{geometricproduct}) is an algebra over $\F{R}$ known as the \textit{geometric algebra} (GA) of $\F{R}^{n}$. Such an algebra is denoted by $\C{G}_{n}$ and has dimension $C(n,0) + C(n,1) + \dots + C(n,n) = 2^{n}$.
\end{defi}
\begin{rem}
	Since the grading structure of multivectors is a property associated with the exterior product, the elements of $\mathcal{G}_{n}$ can still be called $k$-blades, $k$-vectors and multivectors.
\end{rem}
An important family of linear operators in $\C{G}_{n}$ are the \textit{grade-$k$ projection operators}, denoted by $\left<\cdot\right>_{k}$ for $0\leq k\leq n$. Applied to an arbitrary multivector $A$, $\left<A\right>_{k}$ projects onto the grade-$k$ components in $A$, i.e., it returns the components of $A$ that can be expressed as a linear combination of $\{e_{i_{1}}\wedge\dots\wedge e_{i_{k}}\}_{1\leq i_{1}<\dots<i_{k}\leq n}$. Obviously, if $A_{k}$ denotes a $k$-vector, then $\left<A_{k}\right>_{k} = A_{k}$.

Using these operators, general multivectors $A\in\C{G}_{n}$ can be expressed as:
\begin{equation}\label{multivectors}
	A = \left<A\right>_{0} + \left<A\right>_{1} + \dots + \left<A\right>_{n}.
\end{equation}
Hence, the set of all $k$-vectors for a given $1\leq k\leq n$ is a vector subspace of $\mathcal{G}_{n}$ denoted by $\left<\mathcal{G}_{n}\right>_{k}$ and spanned by $B_{k} = \{e_{i_{1}}\wedge\dots\wedge e_{i_{k}}\}_{1\leq i_{1}<\dots<i_{k}\leq n}$.

The multivector representation (\ref{multivectors}) is very useful in defining another important operator in $\C{G}_{n}$. This linear operator is known as the \textit{reversion operator} and is denoted by the superscript $\sim$. The reversion is defined over the geometric product of $m$ vectors as:
\begin{equation}\label{reverse}
	(\bm{a}_{1}\cdots \bm{a}_{m})^{\sim} = \bm{a}_{m}\cdots \bm{a}_{1}.
\end{equation}
Applied to $k$-vectors, we have that:
\begin{equation}\label{reverseblades}
	\widetilde{A}_k = (-1)^{\frac{k(k-1)}{2}}A_{k}
\end{equation}
due to the anticommutativity of the exterior product. Finally, since reversion is a linear operator, the reverse of an arbitrary multivector is:
\begin{equation}\label{reversemultivector}
	\widetilde{A} = \left<\widetilde{A}\right>_{0} + \dots + \left<\widetilde{A}\right>_{n} = \left<A\right>_{0} + \left<A\right>_{1} - \left<A\right>_{2} + \dots + (-1)^{\frac{n(n-1)}{2}}\left<A\right>_{n}.
\end{equation}
Finally, another operator of great interest is the \textit{dual operator}. Every grade-$n$ element of $\C{G}_{n}$ is of the form $\alpha(e_{1}\wedge\dots\wedge e_{n})$ for a scalar $\alpha\in\F{R}$. For each $\alpha\in\F{R}$, $\alpha(e_{1}\wedge\dots\wedge e_{n})$ is known as the \textit{volume element} $E_{\alpha}$ of $\mathcal{G}_{n}$, while the generator $e_{1}\wedge\dots\wedge e_{n}$  is known as the \textit{pseudoscalar} of $\C{G}_{n}$ and is usually denoted by $I$. Pseudoscalars allow us to define the dual operator, whose action over a $k$-vector $A_{k}$ is:
\begin{equation}\label{dual}
	A_{k}^{\ast} = IA_{k},
\end{equation}
where $A_{k}^{\ast}$ is an $(n-k)$-vector.

Now, let us go back to the bivectors of $\mathcal{G}_n$ since they will be fundamental in the modelling of the rotations in $\mathbb{R}^n$. An important property of these bivectors is that they always square to a scalar. Therefore, given a bivector $B$, the unit bivector associated with $B$ is $B' = B/|B^2|$. Unit bivectors of $\mathcal{G}_n$ always square to -1. This allows us to compute the following series:
\begin{equation}\label{expbivector}
	\exp(\alpha B') = \sum\limits_{m=0}^\infty \dfrac{(\alpha B')^m}{m!},
\end{equation}
where $\alpha\in\mathbb{R}$. Expanding equation \eqref{expbivector}, we have that:
\begin{equation}\label{expbivector_expanded}
	\begin{split}
	\exp(\alpha B') &= 1 + \alpha B' - \dfrac{\alpha^2}{2} - \dfrac{\alpha^3B}{3!} + \cdots\\
	&= \left(1 - \dfrac{\alpha^2}{2} + \cdots\right) + B'\left(\alpha- \dfrac{\alpha^3}{3!} + \cdots\right)\\
	&=\cos(\alpha) + B'\sin(\alpha).
	\end{split}
\end{equation}
Equation \eqref{expbivector_expanded} indicates that $\exp(\alpha B')$ could be related to rotations. Indeed, we have the following result.

\begin{pro}
	Let $B$ be a unit bivector and $0\leq \theta\leq 2\pi$, then $R = \exp(-(\theta B)/2) = \cos(\theta/2) - B\sin(\theta/2)\in\left<\mathcal{G}_n\right>_0 + \left<\mathcal{G}_n\right>_2$ defines a rotation by an angle $\theta$ and with rotation plane represented by $B$. It acts over an element $X\in\mathcal{G}_n$ through the \textit{sandwiching product}:
	\begin{equation}
		X' = RX\widetilde{R}.
	\end{equation}
	Such an element $R$ is termed a \textit{rotor}.
\end{pro}
Rotors satisfy the following properties:
\begin{itemize}
	\item[1)] $R\widetilde{R} = 1$.
	\item[2)] $R\bm{x}\widetilde{R} = (-R)\bm{x}(-\widetilde{R})$ for $\bm{x}\in\mathbb{R}^n$.
	\item[3)] $R\bm{x}\bm{y}\widetilde{R} = R\bm{x}\widetilde{R}R\bm{y}\widetilde{R}$ for $\bm{x},\bm{y}\in\mathbb{R}^n$.
\end{itemize} 
The first property is the analogous version of the property defining the orthogonal matrices with determinant equal to 1, which are known to represent rotations. The second property proves that both $R$ and $-R$ encode the same rotation, while the third property is known as the geometric covariance of rotors.

In general, rotors define a group $\mathfrak{R}$ with the geometric product as the group product:
\begin{equation}
	\mathfrak{R} = \{R\in\left<\mathcal{G}_n\right>_0 + \left<\mathcal{G}_n\right>_2\;:\;R\widetilde{R} = 1\}.
\end{equation}
Therefore, the product of two different rotors $R_1$ and $R_2$ also encodes a rotation. In particular, it is the rotation resulting from the composition of the rotations encoded by $R_1$ and $R_2$ respectively. In addition, the second property states that $\mathfrak{R}$ provides a double covering of the rotation group.

Finally, one of the most important geometric algebras is the spatial geometric algebra $\mathcal{G}_{3}$, whose basis is:
\begin{equation}
	\{1,e_{1},e_{2},e_{3},e_{12},e_{13},e_{23},I\},
\end{equation}
where $\{e_{1},e_{2},e_{3}\}$ is an orthonormal basis of $\mathbb{R}^3$ and $ e_{ij} = e_{i}\wedge e_{j}$.

\section{Identification of singularities using geometric algebra}\label{IdentificationSing}
Since the degrees of freedom required to describe the position and orientation of a rigid body in the three-dimensional space are six, the more natural way of formulating the singularity problem is through the six-dimensional geometric algebra $\mathcal{G}_{6}$, that extends naturally the three-dimensional algebra $\mathcal{G}_{3}$ introduced in section \ref{Background}. Screw theory \cite{MurrayLiShankarSastry94,DavidsonHunt04} provides an intuitive and geometrical description of the differential kinematics of serial and parallel manipulators using six-dimensional vectors. Because of this, throughout this chapter some concepts taken from this theory will be employed. This will provide the initial framework to completely understand the approach introduced in this section. 

As stated in the introduction, we are going to work with three-dimensional rigid motions, i.e., three-dimensional orientation-preserving isometries. They form a Lie group, called the \textit{special Euclidean group}, denoted by $SE(3)$. Its associated Lie algebra is:
\begin{equation}
	\mathfrak{se}(3) = \left\{\hat{\xi}\in\mathcal{M}_4\,:\,\hat{\xi}=\begin{pmatrix}
		\Omega & \bm{v}\\
		0 & 0
	\end{pmatrix}\right\},
\end{equation}
where $\Omega$ is a skew-symmetric matrix of order 3, $\bm{v}\in\mathbb{R}^3$ and $\mathcal{M}_4$ denotes the vector space of order 4 square matrices with real entries. Since every skew-symmetric matrix $\Omega$ can be represented as a vector $\bm{\omega}$, we can express an element $\hat{\xi}\in\mathfrak{se}(3)$ as a six-dimensional vector $\xi = [\bm{\omega}\;\;\bm{v}]^T$, termed a \textit{twist}. Therefore, twists are the infinitesimal generators of rigid motions via the exponential map, i.e., $\exp(\hat{\xi}t)=f(t)$ with $f\in SE(3)$. The next theorem is a fundamental result in screw theory.

\begin{theo}[Chasles, 1830]
	Every rigid motion $f\in SE(3)$ can be realized as a rotation around an axis followed (preceded) by a translation along the same axis.
\end{theo}

\begin{defi}
	A \textit{screw motion} consists of a rotation around an axis followed (preceded) by a translation along the same axis, the \textit{screw axis} $\ell$. The ratio between the translational and the rotational part of the motion is known as the \textit{pitch} and denoted by $h$. In particular, if a point is rotated around $\ell$ by an angle $\theta\neq 0$ and translated along $\ell$ an amount $d$, then $h = d/\theta$. By convention, if $\theta = 0$, $h = \infty$.
\end{defi}
\begin{rem}\label{pitch}
	For infinitesimal motions, if $\theta\neq 0$, then the pitch is defined as $h = \dot{d}/\dot{\theta}$.
\end{rem}
Hence, every rigid motion is a screw motion. Particular cases of screw motions are the pure rotations (pure translations) where the translation (rotation) is the identity or, equivalently, $h=0$ ($h=\infty$). In addition, every screw motion can be characterized by the triple $(\ell, h, q)$, where $q$ denotes the \textit{magnitude} of the motion. If $h \neq \infty$, then $\theta = q$ and $d = h\theta$, while if $h = \infty$, then $\theta =0$ and $d = q$. We call this triple the \textit{screw} associated with the screw motion and we denoted it by $\$$.

\begin{pro}\label{pro_twists_screws}
	Given a screw $\$ = (\ell, h, q)$ with screw axis $\ell$, pitch $h$ and magnitude $q$, there exists a twist $\xi$ such that the rigid motion it generates is the screw motion associated with $\$$.
\end{pro}
Proposition \ref{pro_twists_screws} states a correspondence between twists and screws that is useful for our purposes. In particular, if $\bm{p}$ is a point on $\ell$ and $\bm{v}$ is its direction unit vector, then $\ell = \{\bm{p} + \bm{v}\lambda :\lambda\in\mathbb{R}\}$ and we have that:
\begin{equation}
	\begin{array}{ll}
		\xi = \theta\begin{bmatrix}
			\bm{v}\\
			\bm{p}\times\bm{v} + h\bm{v}
		\end{bmatrix}&\text{for a general screw motion,}\\
		\empty & \empty\\
		\xi = \theta\begin{bmatrix}
			\bm{v}\\
			\bm{p}\times\bm{v}
		\end{bmatrix}&\text{for a pure rotation,}\\
		\empty & \empty\\
		\xi = d\begin{bmatrix}
			\bm{0}\\
			\bm{v}
		\end{bmatrix}&\text{for a pure translation.}
	\end{array}
\end{equation}
A twist $\xi$ associated with a magnitude 1 screw $\$$ is said to be a \textit{unit twist}. Hence, any twist $\xi$ can be seen as a unit twist multiplied by the magnitude of the associated screw axis:
\begin{equation}
	\xi = \theta\xi_U = \theta\begin{bmatrix}
		\bm{v}\\
		\bm{p}\times\bm{v} + h\bm{v}
	\end{bmatrix},
\end{equation}
where $\xi_U$ is a unit twist. Clearly, $\xi$ is associated with $\$=(\ell,h,q)$, while $\xi_U$ is associated with $\$ = (\ell,h,1)$.
\begin{pro}
	Let us consider a rigid body performing a screw motion represented by the screw $\$ = (\ell,h,q(t))$, where the magnitude $q(t)$ is a time-dependent variable. Its velocity during the screw motion is given by the associated twist $\xi$ where, now, the pitch is defined as in remark \ref{pitch}. In particular:
	\begin{equation}
		\begin{array}{ll}
			\xi = \dot{\theta}(t)\begin{bmatrix}
				\bm{v}\\
				\bm{p}\times\bm{v} + h\bm{v}
			\end{bmatrix} &\text{if }\theta\neq 0,\\
			\empty & \empty\\
			\xi = \dot{d}(t)\begin{bmatrix}
				\bm{0}\\
				\bm{v}
			\end{bmatrix} &\text{if }\theta = 0,
		\end{array}
	\end{equation}
	where, here, $\dot{\theta}(t)$ ($\dot{d}(t)$) is known as the \textit{twist amplitude}. 
\end{pro}
Now, let us consider a serial robot with $n$ DoF where $\bm{\omega},\bm{v}$ denote the angular and linear velocity vectors of its end-effector. If equation (\ref{kinematicrelation}) is expanded, the following is obtained:
\begin{equation}\label{geometricJacobianexpanded}
	\left[\begin{array}{c}
		\bm{v}\\
		\bm{\omega}
	\end{array}\right] = J_{1}(\bm{q})\dot{q}_{1} + \dots + J_{n}(\bm{q})\dot{q}_{n},
\end{equation}
where $J_{i}$ denotes the $i$-th column of the Jacobian matrix $J$. Notice that the right side of equation (\ref{geometricJacobianexpanded}) can be seen as the addition of the twists associated with the joints of the robot, where $\dot{q}_i$ plays the role of the twist amplitude and where the linear and angular parts are interchanged. However, for the sake of formality, let us consider the unit twist $\xi_i$ associated with the $i$-th joint of the robot (since, from now on, we are going to work exclusively with unit twists, the subindex $U$ is omitted for simplicity). Then:
\begin{equation}\label{twist_jacobian}
	\xi_{i}(\bm{q})\dot{q}_i=\left\{\begin{array}{lcl}
		\left[\begin{array}{c}
			\bm{z}_{i}\\
			\bm{z}_{i}\times(\bm{o}_{n}-\bm{o}_{i})
		\end{array}\right]\dot{q}_{i}&\quad\quad&\text{if joint }i\text{ is revolute}\\
		\left[\begin{array}{c}
			\bm{0}\\
			\bm{z}_{i}
		\end{array}\right]\dot{q}_{i}&\quad\quad&\text{if joint }i\text{ is prismatic}
	\end{array}\right.
\end{equation}
where, as stated in the introduction, $\bm{z}_{i}$ is the direction vector of the joint axis, $\bm{o}_{n}$ ($\bm{o}_{i}$) is the origin of the frame attached to the end-effector ($i$-th joint) and $\dot{q}_i = \dot{\theta}_i$ if joint $i$ is revolute and $\dot{q}_i = \dot{d}_i$ if joint $i$ is prismatic. 

\begin{rem}
	The unit twists $\xi_i(\bm{q})$ defined in equation \eqref{twist_jacobian} are represented with respect to the world frame, not with respect to the local frame attached to the previous joint. If the unit twists are defined with respect to a local frame, we need to use the adjoint transformation to represent them with respect to the world frame. In particular, $\xi'_i(\bm{q}) = \text{Ad}_f\xi_i(\bm{q})$, where $\text{Ad}_f:\mathbb{R}^6\to\mathbb{R}^6$ is the adjoint transformation associated with the rigid motion $f$, i.e., the rigid motion transforming the reference frame to the local frame in which the twist is initially represented.
\end{rem}
The following is a key result:
\begin{theo}[Tsai, 1999 \cite{Tsai99}]
	Given a serial robot with $n$ DoF:
	\begin{equation}\label{ScrewJacobian}
		\left[\begin{array}{c}
			\bm{\omega}\\
			\bm{v}
		\end{array}\right] = \xi_{1}(\bm{q})\dot{q}_{1} + \dots + \xi_{n}(\bm{q})\dot{q}_{n}  = [\xi_{1}(\bm{q})\;\cdots\;\xi_{n}(\bm{q})]\dot{\bm{q}}, 
	\end{equation}
	where, again, $\bm{\omega},\bm{v}$ denote the angular and linear velocity vectors of the robot's end-effector and $\dot{\bm{q}} = (\dot{q}_{1},\dots,\dot{q}_{n})$.
\end{theo}
The main advantage of the screw-based Jacobian matrix defined in equation (\ref{ScrewJacobian}) is that it allows a geometrical identification of the singularities. Moreover, if an approach based on geometric algebra is used, an intuitive geometrical and computer-friendly algebraic identification of the singularities is possible. For that purpose, let us consider the geometric algebra $\mathcal{G}_{6}$ where for every $i=1,\dots,n$, the unit twist $\xi_{i}(\bm{q})$ can be modelled as a vector. Indeed, we make the identification $\xi_i(\bm{q}) = [\xi_{i_{1}}\cdots\;\xi_{i_{6}}]^T$ with the vector $x = \xi_{i_{1}}e_1 + \cdots + \xi_{i_{6}}e_6\in\mathcal{G}_{6}$, where $e_{1},\dots,e_{6}$ are the basis vectors of $\mathcal{G}_{6}$.

The following gives the main result of this section.
\begin{theo}\label{DetG6}
	Let $\xi_{i}(\bm{q})$ denote the unit twist defined by the $i$-th joint expressed as a vector of $\mathcal{G}_{6}$. Then:
	\begin{equation}\label{DetG6Main}
		\xi_{1}(\bm{q})\wedge\cdots\wedge \xi_{6}(\bm{q}) = \det([\xi_{1}(\bm{q})\; \cdots \; \xi_{6}(\bm{q})])e_{1}\wedge\cdots\wedge e_{6}.
	\end{equation}
\end{theo}
Theorem \ref{DetG6} can be seen as a particular case of a more general result:
\begin{theo}\label{DetGn}
	Let $\bm{a}_{1},\dots,\bm{a}_{n}$ be a set of $n$ vectors of $\mathcal{G}_{n}$. Then:
	\begin{equation}
		\bm{a}_{1}\wedge\cdots\wedge \bm{a}_{n} = \det([\bm{a}_{1}\; \cdots \; \bm{a}_{n}])e_{1}\wedge\cdots\wedge e_{n}
	\end{equation}
\end{theo}
\begin{proof}
	We define a linear transformation $F:\mathbb{R}^n\to\mathbb{R}^n$ that we extend to a linear transformation in $\mathcal{G}_n$, $F:\mathcal{G}_n\to\mathcal{G}_n$, by asking $F$ to satisfy that $F(\bm{a}\wedge\cdots\wedge\bm{b}) = F(\bm{a})\wedge 
	\cdots\wedge F(\bm{b})$ for any set of vectors. We define $F$ as follows:
	\begin{equation}
		\begin{split}
			F(e_1) &= \bm{a}_1\\
			&\vdots\\
			F(e_n) &= \bm{a}_n\\
		\end{split}
	\end{equation}
	Hence, $F$ transforms the set of vectors $\{e_1,\dots,e_n\}$ into $\{\bm{a}_1,\dots,\bm{a}_n\}$ and, thus, its associated matrix is:
	\begin{equation}
		M_F = \begin{pmatrix}
			\horzbar & \bm{a}_1 & \horzbar\\
			\horzbar & \bm{a}_2 & \horzbar\\
			\empty & \vdots & \empty\\
			\horzbar & \bm{a}_n & \horzbar
		\end{pmatrix}.
	\end{equation}
	Now, we have the following result \cite[pag. 108]{DoranLasenby03}:
	\begin{equation}
		F(I) = \det(F)I,
	\end{equation}	
	where, as stated in section \ref{Background}, $I$ denotes the pseudoscalar of $\mathcal{G}_n$. Now, the result can be easily derived since:
	\begin{equation}
		\bm{a}_1\wedge \cdots\wedge\bm{a}_n = F(e_1)\wedge\cdots\wedge F(e_n) = F(e_1\wedge\cdots\wedge e_n) = \det(M_F)e_1\wedge\cdots\wedge e_n,
	\end{equation}
	where, clearly, $\det(M_F) = \det(\bm{a}_1\cdots \bm{a}_n)$.
\end{proof}
In particular, theorem \ref{DetGn} is true for any set of six vectors $\bm{a}_{1},\dots,\bm{a}_{6}$ of $\mathcal{G}_{6}$, which proves theorem \ref{DetG6}. Now, the following corollary of theorem \ref{DetG6} allows us to characterize the singularities of any serial robot of 6 DoF.
\begin{cor}\label{caracterizacionsingularidades}
	Given a serial robot with 6 DoF and associated unit twists $\xi_{1}(\bm{q}),\dots,\xi_{6}(\bm{q})$, then $\bm{q}\in\mathcal{S}$ if, and only if, $\xi_{1}(\bm{q})\wedge\dots\wedge \xi_{6}(\bm{q}) = 0$.
\end{cor}
\begin{proof}
	Taking the dual of equation (\ref{DetG6Main}), the following identity is obtained:
	\begin{equation}\label{DetG6Dual}
		(\xi_{1}(\bm{q})\wedge\cdots\wedge\xi_{6}(\bm{q}))^{\ast} = \det([\xi_{1}(\bm{q})\; \cdots \; \xi_{6}(\bm{q})])
	\end{equation}
	and, therefore, the singularities of the serial robot are those configurations $\bm{q}\in\mathcal{C}$ verifying that:
	\begin{equation}\label{eqSing}
		(\xi_{1}(\bm{q})\wedge\cdots\wedge \xi_{6}(\bm{q}))^{\ast} = 0.
	\end{equation}
	Now, since for a given non-zero multivector $M\in\mathcal{G}_{n}$, $M^{\ast} = 0$ if, and only if, $M = 0$, equation (\ref{eqSing}) can be simplified to:
	\begin{equation}\label{eqSingSimplified}
		\xi_{1}(\bm{q})\wedge\cdots\wedge \xi_{6}(\bm{q}) = 0.
	\end{equation}
	Thus, $\bm{q}\in\mathcal{S}$ if, and only if, $\xi_{1}(\bm{q})\wedge\cdots\wedge \xi_{6}(\bm{q}) = 0$.
\end{proof}
In addition, corollary \ref{caracterizacionsingularidades} allows us to re-define the singular set as:
\begin{equation}\label{singularset}
	\mathcal{S} = \{\bm{q}\in\mathcal{C}\,:\,\xi_{1}(\bm{q})\wedge\cdots\wedge\xi_{6}(\bm{q}) = 0\}.
\end{equation}

\begin{rem}\label{remequalscrews}
	What theorem \ref{DetG6} states is that, for instance, if two unit twists $\xi_{1}$ and $\xi_{2}$ satisfy $\xi_{1}\wedge\xi_{2}= 0$, then they represent the same twist, and hence, they generate the same screw motion. This means that, if such a screw motion is a pure translation, then the translational axes are either parallel or coincident, while if the screw motion is a pure rotation, the rotational axes are coincident (since the twists contains the term $(\bm{z}_{i}\times(\bm{o}_{6}-\bm{o}_{i}))$ for $i = 1,2$, they cannot be parallel). Regarding the kinematic singularities of serial robots, this implies that two prismatic joints whose axes are either parallel or coincident give rise to a singularity and, equivalently, that two revolute joints whose axes are coincident give rise to a singularity. This is, in fact, in agreement with what it is known about kinematic singularities since two parallel revolute joint axes do not give rise to a singularity. Obviously, the same geometrical interpretation can be made for three, four or more unit twists satisfying that their outer product is zero.
\end{rem}

With respect to redundant serial robots, it is clear that, for $n>6$, $\xi_{1}(\bm{q})\wedge\dots\wedge\xi_{n}(\bm{q}) = 0$ for any $\bm{q}\in\mathcal{C}$. Hence, corollary \ref{caracterizacionsingularidades} by its own does not allow us to characterize the singularities of redundant robots. However, this problem can be easily overcome by studying all the possible combinations of six unit twists in $\{\xi_{1}(\bm{q}),\dots,\xi_{n}(\bm{q})\}$. We denote the set of all combinations of six elements that can be drawn from $\{1,\dots,n\}$ by $S$. Clearly, $S$ has $C(n,6)=\binom{n}{6}$ elements of the form $\{i_1,\dots,i_6\}$, where $1\leq i_1<\cdots<i_6\leq n$ and $1\leq i\leq C(n,6)$.
\begin{theo}\label{caracterizationredundante}
	Given a serial robot with $n$ DoF and associated unit twists $\xi_{1}(\bm{q}),\dots,\xi_{n}(\bm{q})$, then $\bm{q}\in\mathcal{S}$ if, and only if, for each $1\leq i\leq C(n,6)$:
	\begin{equation}\label{DetG6redundant}
		\xi_{i_{1}}(\bm{q})\wedge\dots\wedge\xi_{i_{6}}(\bm{q}) = 0,
	\end{equation}
	where $\{i_1,\dots,i_6\}$ is the $i$-th element of $S$.
\end{theo}
\begin{proof}
	It follows that, for $\bm{q}\in\mathcal{C}$:
	\begin{equation}\label{MenoresChapter6}
		\begin{split}
			&\xi_{i_{1}}(\bm{q})\wedge\dots\wedge \xi_{i_{6}}(\bm{q}) = 0\text{ for every }1\leq i\leq C(n,6)\\
			&\stackrel{(1)}{\Longleftrightarrow}\det([\xi_{i_{1}}(\bm{q})\;\cdots\;\xi_{i_{6}}(\bm{q})])=0\text{ for every }1\leq i\leq C(n,6)\\
			&\stackrel{(2)}{\Longleftrightarrow}\rho([\xi_{1}(\bm{q})\;\cdots\;\xi_{n}(\bm{q})])<6
		\end{split}
	\end{equation}
	where $(1)$ uses equation (\ref{DetG6Dual}) and $(2)$ uses the fact that all the minors of order 6 of the matrix $[\xi_{1}(\bm{q})\;\cdots\;\xi_{n}(\bm{q})]$ have null determinant. Clearly, $\rho([\xi_{1}(\bm{q})\;\cdots\;\xi_{n}(\bm{q})])<6$ if, and only if, $\rho(J(\bm{q}))<6$ which, in turn, is equivalent to $\bm{q}\in\mathcal{S}$ (by definition \ref{sing_def}).
\end{proof}
The computation of either equation (\ref{eqSingSimplified}) for non-redundant robots or equation (\ref{DetG6redundant}) for redundant ones is computationally more efficient than the computation of either $\det{J(\bm{q})} = 0$ or $\det(J(\bm{q})J^{T}(\bm{q}))= 0$. The main reason for this lies in the computational complexity of the operations needed to obtain the expressions (\ref{eqSingSimplified}) or (\ref{DetG6redundant}) with respect to the complexity of the operations needed for obtaining $\det{J(\bm{q})} = 0$ or $\det(J(\bm{q})J^{T}(\bm{q}))= 0$. It is clear that the outer product of $n$ vectors of $\mathcal{G}_{n}$ behaves like the addition and product of real numbers and, hence, it has complexity $O(n)+O(n^{2})$, while the determinant has complexity $O(n^{3})$ or $O(n^{4})$ depending on the algorithm used. In addition, for redundant robots, there are two main operations: the product between $J(\bm{q})$ and $J^{T}(\bm{q})$ and the determinant of the product matrix. This implies that, for this case, the complexity increases to $O(n^{3})+O(n^{4})$.

\begin{figure}[t!]
	\centering
	\includegraphics[width=10cm]{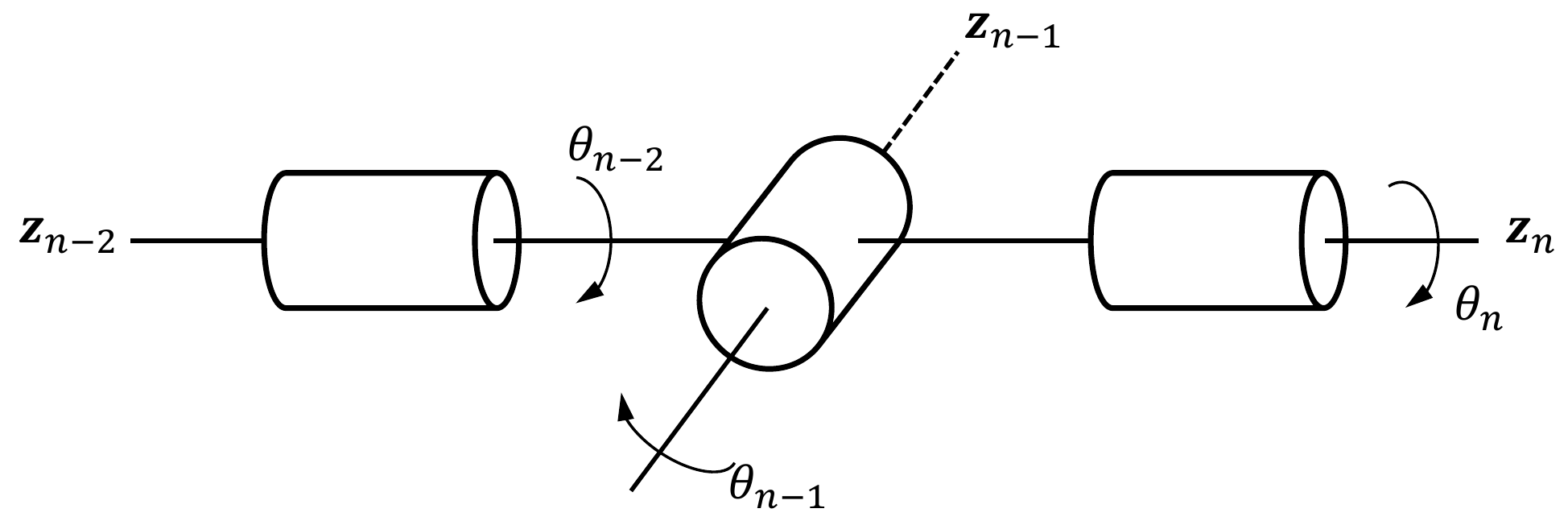}
	\caption{Schematic representation of the wrist singularity.}
	\label{wristsingularity}
\end{figure}

\subsection{Special case: serial robots with a spherical wrist}\label{SpecialCase}
Similarly to what happens with the Jacobian matrix $J$, a simplification can be achieved for robots that have a spherical wrist. As stated in section \ref{Intro}, the singularities of these robots can be decoupled into position and orientation singularities. Position singularities involve the first $n-3$ joints and are computed by studying the rank of the following matrix:
\begin{equation}
	J_{p} = \left[\begin{array}{ccc}
		\bm{s}_{1} & \cdots & \bm{s}_{n-3}
	\end{array}\right],
\end{equation}
where $\bm{s}_i = \bm{z}_{i}\times(\bm{o}_{n}-\bm{o}_{i})$ if joint $i$ is revolute and $\bm{s}_i = \bm{z}_{i}$ if joint $i$ is prismatic. On the other hand, orientation singularities involves the last three joints and are computed through the determinant of the following matrix:
\begin{equation}
	J_{o} = \left[\begin{array}{ccc}
		\bm{z}_{n-2} & \bm{z}_{n-1} & \bm{z}_{n}
	\end{array}\right].
\end{equation}
Now, let us consider the three-dimensional geometric algebra $\mathcal{G}_{3}$. As proven in theorem \ref{DetGn} for $n=3$, $\bm{a}_{1}\wedge \bm{a}_{2}\wedge \bm{a}_{3} = \det([\bm{a}_{1}\;\bm{a}_{2}\;\bm{a}_{3}])e_{1}\wedge e_{2}\wedge e_{3}$ for any three vectors $\bm{a}_{1},\bm{a}_{2},\bm{a}_{3}\in\mathbb{R}^{3}$. Hence, analogously to what has been done before, the following characterization for the position and orientation singularities can be deduced.
\begin{theo}\label{theoremsimplifiedsingularities}
	Given a serial robot with $n$ DoF and a spherical wrist, if either $\bm{z}_{i}\times(\bm{o}_{n}-\bm{o}_{i})$ or $\bm{z}_{i}$ are denoted by $\bm{s}_{i}$ for $i=1,\dots,n-3$, then:
	\begin{itemize}
		\item $\bm{q}\in\mathcal{C}$ is a position singularity if, and only if, $\bm{s}_{i_{1}}(\bm{q})\wedge \bm{s}_{i_{2}}(\bm{q})\wedge \bm{s}_{i_{3}}(\bm{q}) = 0$ for each $1\leq i\leq C(n-3,3)$, where $\{i_1,i_2,i_3\}$ is the $i$-th combination of three elements drawn from $\{1,\dots,n-3\}$.
		\item $\bm{q}\in\mathcal{C}$ is an orientation singularity if, and only if, 
		\begin{equation}\label{wrist_sing}
			\bm{z}_{n-2}(\bm{q})\wedge\bm{z}_{n-1}(\bm{q})\wedge\bm{z}_{n}(\bm{q}) = 0.
		\end{equation}
	\end{itemize}
\end{theo}
\begin{proof}
	The proof is completely analogous to the proof of corollary \ref{caracterizacionsingularidades} and theorem \ref{caracterizationredundante}.
\end{proof}
\begin{rem}\label{remwristsingularity}
	Since the last three joint axes either intersect at a single point or are parallel, there is only one orientation singularity, namely when these three joint axes are coplanar. This can also be easily deduced from equation \eqref{wrist_sing}. A schematic representation of such singularity, also called \textit{wrist singularity}, is depicted in figure \ref{wristsingularity}.
\end{rem}

\section{Distance to singularities}\label{Distance}
\begin{figure}[t!]
	\centering
	\includegraphics[width=12cm,height=8cm]{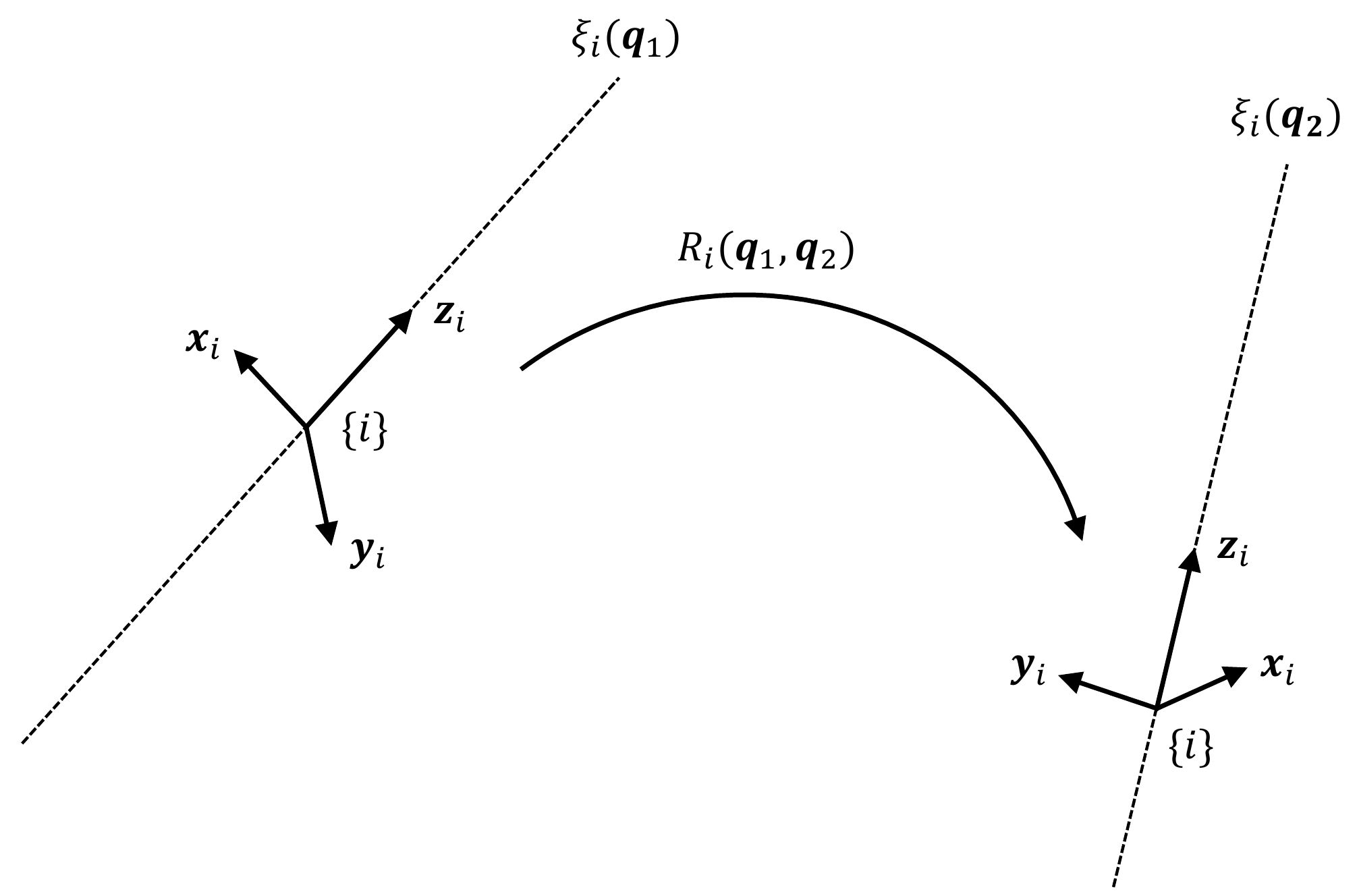}
	\caption{Rotor $R_{i}$ relating the twist $\xi_{i}$ in two different configurations $\bm{q}_{1}$ and $\bm{q}_{2}$.}
	\label{RotorConfiguration}
\end{figure}

Let $\bm{q}_{1},\bm{q}_{2}\in\mathcal{C}$ be two arbitrary configurations of a serial robot with $n$ DoF and let $\xi_{1},\dots,\xi_{n}$ be the unit twists associated with its joints. Then, there exist $R_{1}(\bm{q}_{1},\bm{q}_{2}),\dots, R_{n}(\bm{q}_{1},\bm{q}_{2})$, where, for each $1\leq i\leq n$, $R_{i}(\bm{q}_{1},\bm{q}_{2})$ is a configuration-dependent rotor in the six-dimensional geometric algebra $\mathcal{G}_{6}$ such that (figure \ref{RotorConfiguration}):
\begin{equation}
	\xi_{i}(\bm{q}_{2}) = R_{i}(\bm{q}_{1},\bm{q}_{2})\xi_{i}(\bm{q}_{1})\widetilde{R}_{i}(\bm{q}_{1},\bm{q}_{2}).
\end{equation}
The reason why these rotors exist is simple: unit twists are modelled as vectors in $\mathcal{G}_6$ and there always exists a rotor relating any pair of vectors in any geometric algebra $\mathcal{G}_n$. In particular, there is always a rotor relating the same unit twist $\xi$ in two different configurations $\bm{q}_{1},\bm{q}_{2}$.

Now, let $\bm{q}_{s}\in\mathcal{S}$ denote a singularity of a serial robot. As explained in the previous section, if the serial robot has a spherical wrist, then $\bm{q}_{s}$ only involves a maximum of two or three joints and, therefore, two or three unit twists. If, conversely, the robot has not spherical wrist, then it can involve a maximum of six joints. Let us suppose, without loss of generality, that a given singularity $\bm{q}_{s}$ involve the joints $i_1,\dots,i_r$ with associated unit twists $\xi_{i_{1}}(\bm{q}_{s}),\dots,\xi_{i_{r}}(\bm{q}_{s})$ for $2\leq r\leq 6$. Then, for any configuration $\bm{q}\in\mathcal{C}$, there exist $R_{i_{1}}(\bm{q},\bm{q}_{s}),\dots,R_{i_{r}}(\bm{q},\bm{q}_{s})$ such that:
\begin{equation}
	\xi_{i_{j}}(\bm{q}_{s}) = R_{i_{j}}(\bm{q},\bm{q}_{s})\xi_{i_{j}}(\bm{q})\widetilde{R}_{i_{j}}(\bm{q},\bm{q}_{s})\text{ for each } 1\leq j\leq r.
\end{equation}
The notation chosen for these rotors expresses a configuration dependence that is not a functional dependency, i.e., there is not an analytical expression for these rotors with $\bm{q}$ as a variable.

Now, it is clear that $R_{i_{j}}(\bm{q},\bm{q}_{s}) = 1$ if, and only if, $\bm{q} = \bm{q}_{s}$ for every $j=1,\dots,r$. However, since for each $j$, $R_{i_{j}}(\bm{q},\bm{q}_{s})$ does not define a function on $\bm{q}$, a distance function cannot be defined. But, the measure of how close is a given configuration $\bm{q}$ to a singularity can be set as:
\begin{equation}
	\bm{q}\approx\bm{q}_{s}\Longleftrightarrow R_{i_{j}}(\bm{q})\approx 1\text{ for every }j=1,\dots,r.
\end{equation}
\vspace{0.2cm}
\begin{ejem}
	Let $\bm{q}_{s}\in\mathcal{S}$ be a singularity of a serial robot that only involves the second and third joints. Then, for any configuration $\bm{q}\in\mathcal{C}$ there exist $R_{2}(\bm{q},\bm{q}_{s})$ and $R_{3}(\bm{q},\bm{q}_{s})$ such that:
	\begin{equation}
		\begin{split}
			\xi_{2}(\bm{q}_{s}) &= R_{2}(\bm{q},\bm{q}_{s})\xi_{2}(\bm{q})\widetilde{R}_{2}(\bm{q},\bm{q}_{s}),\\
			\xi_{3}(\bm{q}_{s}) &= R_{3}(\bm{q},\bm{q}_{s})\xi_{3}(\bm{q})\widetilde{R}_{3}(\bm{q},\bm{q}_{s}).
		\end{split}
	\end{equation}
	Therefore, $\bm{q}$ is close to $\bm{q}_{s}$ if, and only if:
	\begin{equation}
		\left.\begin{split}
			R_{2}(\bm{q},\bm{q}_{s})&\approx 1\\
			R_{3}(\bm{q},\bm{q}_{s})&\approx 1
		\end{split}\right\}
	\end{equation}
	and is singular if, and only if:
	\begin{equation}
		\left.\begin{split}
			R_{2}(\bm{q},\bm{q}_{s})&= 1\\
			R_{3}(\bm{q},\bm{q}_{s})&= 1
		\end{split}\right\}
	\end{equation}
	where, in general, $R_{2}(\bm{q},\bm{q}_{s})\neq R_{3}(\bm{q},\bm{q}_{s})$.
\end{ejem}
These rotors can be constructed in many different ways. The easiest way consists of considering, for each $1\leq i\leq n$, the frame $\{i\}$ attached to joint $i$ and constructed from $\xi_i$. This three-dimensional frame varies with the configuration $\bm{q}$. Hence, for two different configurations $\bm{q}_1$ and $\bm{q}_2$, there are two frames $\{i\}$ attached to joint $i$. As shown in \cite{DoranLasenby03}, we can recover the three-dimensional rotor that transforms one of the frames into the other. Since each frame $\{i\}$ depends continuously on the configuration $\bm{q}$, the rotor $R_{i}(\bm{q})$ is a continuous function defined as follows:
\begin{equation}
	\begin{split}
		R_{i}:\mathcal{C}&\to\mathfrak{R}\\
		\bm{q}&\mapsto R_{i}(\bm{q})
	\end{split}
\end{equation}
Thus, these configuration-dependent rotors exhibit a functional dependency on the configuration, which allows us to define a distance function. Such distance is based on the norm of a multivector $X\in\mathcal{G}_{n}$, defined by the relation:
\begin{equation}
	\|X\|^2 = \left<X\widetilde{X}\right>_{0}.
\end{equation}
To prove that $\|\cdot\|$ is a norm, the following two lemmas are necessary. 
\begin{lem}\label{lemmanorma}
	For any given multivector $X\in\mathcal{G}_{n}$, $\left<X\widetilde{X}\right>_{0}\in\F{R}^{+} = [0,+\infty)$.	
\end{lem}
\begin{proof}
	According to equation (\ref{reversemultivector}), it follows that:
	\begin{equation}
		\widetilde{X} = \left<\widetilde{X}\right>_{0} + \left<\widetilde{X}\right>_{1} + \dots + \left<\widetilde{X}\right>_{n}.
	\end{equation}
	Then:
	\begin{equation}\label{sumdistributive}
		X\widetilde{X} = \sum_{i=0}^{n} \sum_{j=0}^{n} \left<X\right>_{i}\left<\widetilde{X}\right>_{j}.
	\end{equation}
	Now, note that, for each $i=1,\dots,n$, $\left<X\right>_{i}$ is a $i$-vector, i.e., it only contains terms of grade $i$. The geometric product of two $k$-vectors (with different $k$) is stated as follows \cite[pp. 103]{DoranLasenby03}:
	\begin{equation}
		A_{r}B_{s} = \left<A_{r}B_{s}\right>_{|r-s|} + \left<A_{r}B_{s}\right>_{|r-s|+2} + \dots + \left<A_{r}B_{s}\right>_{r+s},
	\end{equation}
	Therefore, it is clear that $\left<\left<X\right>_{i}\left<\widetilde{X}\right>_{j}\right>_{0} = 0$ for $i\neq j$. Thus:
	\begin{equation}\label{decompositionmultivectorchapter6}
		\left<X\widetilde{X}\right> _{0} = \sum_{i=0}^{n} \left<\left<X\right>_{i}\left<\widetilde{X}\right>_{i}\right>_{0}.
	\end{equation}
	Now, each $\left<X\right>_{i}$ can be expanded as follows:
	\begin{equation}
		\left<X\right>_{i} = \sum_{j=1}^{C(n,i)}\alpha_{j}(i)e_{j_{1}}\cdots e_{j_{i}},
	\end{equation}
	where, for every $1\leq j\leq C(n,i) = \binom{n}{i}$, $\alpha_{j}(i)\in\mathbb{R}$ and $e_{j_{1}}\cdots e_{j_{i}}$ are the basis elements of $\left<\mathcal{G}_{n}\right>_{i}$. Therefore:
	\begin{equation}
		\left<\widetilde{X}\right>_{i} = \sum_{j=1}^{C(n,i)}\alpha_{j}(i)e_{j_{i}}\cdots e_{j_{1}}
	\end{equation}
	and, thus:
	\begin{equation}
		\begin{split}
			\left<X\right>_{i}\left<\widetilde{X}\right>_{i} &= \sum_{j=1}^{C(n,i)}\alpha_{j}(i)e_{j_{1}}\cdots e_{j_{i}}\sum_{j=1}^{C(n,i)}\alpha_{j}(i)e_{j_{i}}\cdots e_{j_{1}}\\
			&= \sum_{j=1}^{C(n,i)}\sum_{k=1}^{C(n,i)}\alpha_{j}(i)\alpha_{k}(i)e_{j_{1}}\cdots e_{j_{i}}e_{k_{i}}\cdots e_{k_{1}},
		\end{split}
	\end{equation}
	where, clearly, $\left<e_{j_{1}}\cdots e_{j_{i}}e_{k_{i}}\cdots e_{k_{1}}\right>_{0} = \delta_{jk}$ with $\delta_{jk}$ the Kronecker delta. Then:
	\begin{equation}
		\left<\left<X\right>_{i}\left<\widetilde{X}\right>_{i}\right>_{0} = \sum_{j=1}^{C(n,i)}\alpha_{j}^{2}
	\end{equation}
	that, for every $1\leq i\leq n$, is a positive scalar. This implies that the sum of equation (\ref{decompositionmultivectorchapter6}) is also a positive scalar. 
\end{proof}
\begin{lem}\label{threepositivescalars}
	Given three strictly positive real numbers $a_{1},a_{2},a_{3}\in\mathbb{R}^{+}\setminus\{0\}$, the following properties hold:
	\begin{itemize}
		\item $\sqrt{a_{1}+a_{2}-a_{3}}\leq\sqrt{a_{1}} + \sqrt{a_{2}}$.
		\item $\sqrt{a_{1}+a_{2}+a_{3}}\leq\sqrt{a_{1}}+\sqrt{a_{2}}\:$ if, and only if, $\:a_{3}\leq 2\sqrt{a_{1}a_{2}}$.
	\end{itemize}
\end{lem}
\begin{proof}
	Both properties can be obtained by a straightforward computation.
	%	 \begin{equation}
	%	 \begin{split}
	%	 	&\sqrt{a_{1}+a_{2}+a_{3}}\leq\sqrt{a_{1}}+\sqrt{a_{2}}\Longleftrightarrow\\
	%	 	&a_{1}+a_{2}+a_{3}\leq a_{1}+a_{2} + 2\sqrt{a_{1}a_{2}}\Longleftrightarrow\\
	%	 	&a_{3}\leq 2\sqrt{a_{1}a_{2}}
	%	 \end{split}
	%	 \end{equation}
\end{proof}

\begin{pro}\label{propositionnorma}
	The function $\|\cdot\|:\mathcal{G}_{n}\to\F{R}^{+}$ defined by the identity $\|X\|^{2} = \left<X\widetilde{X}\right>_{0}$ is a norm in $\mathcal{G}_{n}$, i.e.,:
	\begin{itemize}
		\item[(i)] $\|X\| \geq 0$ for all $X\in\mathcal{G}_{n}$. In particular, $\|X\| = 0$ if, and only if, $X=0$.
		\item[(ii)] $\|\lambda X\| = |\lambda|\|X\|$ for all $X\in\mathcal{G}_{n}$ and $\lambda\in\mathbb{R}$.
		\item[(iii)] $\|X + Y\| \leq \|X\| + \|Y\|$ for all  $X,Y\in\mathcal{G}_{n}$ (usually known as the \textit{triangle inequality}).
	\end{itemize}
\end{pro}
\begin{proof} \
	\begin{itemize}
		\item[(i)] Given a multivector $X$, identity $\|X\|^{2} = \left<X\widetilde{X}\right>_{0}$ is equivalent to:
		\begin{equation}\label{raiznorma}
			\|X\| = \pm\sqrt{\left<X\widetilde{X}\right>_{0}}.
		\end{equation} 
		Thus, it is clear by lemma \ref{lemmanorma} that the positive branch of equation (\ref{raiznorma}) is well defined and that $\|X\| \geq 0$. In particular, if $\|X\| = 0$, then:
		\begin{equation}
			\sqrt{\left<X\widetilde{X}\right>_{0}} = 0\Longrightarrow \left<X\widetilde{X}\right>_{0} = 0\Longrightarrow \sum_{i=0}^{n} \left<\left<X\right>_{i}\left<\widetilde{X}\right>_{i}\right>_{0} = 0,
		\end{equation}
		where all the terms of the last equation are positive by lemma \ref{lemmanorma} and, thus, all of them are equal to zero. Now, note that each addend is the geometric product of an $i$-vector with its reverse. Therefore, if such product is zero, the corresponding $i$-vector must be zero. Since all the terms are zero, all the $i$-vectors that form $X$ are zero and, thus, $X$ is zero.
		\item[(ii)] If $\lambda\in\mathbb{R}$ and $X\in\mathcal{G}_{n}$, then:
		\begin{equation}
			\begin{split}
				\|\lambda X\| &= \sqrt{\left<(\lambda X)(\lambda\widetilde{X})\right>_{0}} = \sqrt{\left<\lambda^{2}X\widetilde{X}\right>_{0}} \\
				&\stackrel{(1)}{=}\sqrt{\lambda^{2}\left<X\widetilde{X}\right>_{0}} = |\lambda|\sqrt{\left<X\widetilde{X}\right>_{0}} = |\lambda|\|X\|,
			\end{split}
		\end{equation}
		where $(1)$ uses the linearity of the grade-0 projection operator (as stated in section \ref{Background}).
		\item[(iii)] Given two different multivectors $X$ and $Y$, they can be expanded as linear combinations of the basis elements of $\mathcal{G}_{n}$ as follows:
		\begin{equation}
			\begin{split}
				X &= \sum_{i=0}^{2^{n}}\alpha_{i}e_{j_{1}}\cdots e_{j_{i}},\\
				Y &= \sum_{i=0}^{2^{n}}\beta_{i}e_{j_{1}}\cdots e_{j_{i}}.
			\end{split}
		\end{equation}
		Now, it follows that:
		\begin{equation}
			X+ Y = \sum_{i=0}^{2^{n}}(\alpha_{i}+\beta_{i})e_{j_{1}}\cdots e_{j_{i}}
		\end{equation}
		and, hence:
		\begin{equation}
			\begin{split}
				\|X+Y\| &= \sqrt{\left<\left(X+Y\right)\left(X+Y\right)^\sim\right>_{0}} \stackrel{(1)}{=} \sqrt{\sum_{i=0}^{2^{n}}(\alpha_{i}+\beta_{i})^{2}}\\
				&= \sqrt{\commentedbox{\sum_{i=0}^{2^{n}}\alpha_{i}^{2}}{A} + \commentedbox{\sum_{i=0}^{2^{n}}\beta_{i}^{2}}{B} + 2\commentedbox{\sum_{i=0}^{2^{n}}\alpha_{i}\beta_{i}}{C}},
			\end{split}
		\end{equation}
		where $(1)$ uses lemma \ref{lemmanorma}, while $A,B$ and $C$ are just a notation given to simplify the different manipulations. Since $A,B> 0$ (if either $A,B$ are equal to zero, then either $X=0$ or $Y=0$, which will make the condition $\|X+Y\|\leq \|X\|+\|Y\|$ trivial):
		\begin{equation}
			\sqrt{A + B + C} \stackrel{(1)}{\leq}\sqrt{A} + \sqrt{B} = \sqrt{\sum_{i=0}^{2^{n}}\alpha_{i}^{2}} + \sqrt{\sum_{i=0}^{2^{n}}\beta_{i}^{2}} = \|X\| + \|Y\|,
		\end{equation}
		where $(1)$ uses the first or second property of lemma \ref{threepositivescalars} depending on whether $C< 0$ or $C> 0$. It only remains to check that, if $C> 0$, $2C\leq2\sqrt{AB}$. Indeed, the previous inequality is equivalent to that $C^{2}\leq AB$. Now:
		\begin{equation}
			\begin{split}
				AB &= \sum_{i=0}^{2^{n}}\alpha_{i}^{2}\sum_{i=0}^{2^{n}}\beta_{i}^{2} = \sum_{i=0}^{2^{n}}\sum_{j=0}^{2^{n}}\alpha_{i}^{2}\beta_{j}^{2} = \sum_{i=0}^{2^{n}}\alpha_{i}^{2}\beta_{i}^{2} + \sum_{i=0}^{2^{n}}\sum_{\substack{j=0\\[1mm]j\neq i}}^{2^{n}}\alpha_{i}^{2}\beta_{j}^{2},\\
				C^{2} &= \left(\sum_{i=0}^{2^{n}}\alpha_{i}\beta_{i}\right)^{2} = \sum_{i=0}^{2^{n}}\alpha_{i}^{2}\beta_{i}^{2} + \sum_{i=0}^{2^{n}}\sum_{\substack{j=0\\[1mm]j\neq i}}^{2^{n}}\alpha_{i}\beta_{i}\alpha_{j}\beta_{j}
			\end{split}
		\end{equation}
		and, thus, $C^{2}\leq AB$ turns to:
		\begin{equation}
			\sum_{i=0}^{2^{n}}\alpha_{i}^{2}\beta_{i}^{2} + \sum_{i=0}^{2^{n}}\sum_{\substack{j=0\\[1mm]j\neq i}}^{2^{n}}\alpha_{i}\beta_{i}\alpha_{j}\beta_{j}\leq\sum_{i=0}^{2^{n}}\alpha_{i}^{2}\beta_{i}^{2} + \sum_{i=0}^{2^{n}}\sum_{\substack{j=0\\[1mm]j\neq i}}^{2^{n}}\alpha_{i}^{2}\beta_{j}^{2},
		\end{equation}
		that is equivalent to:
		\begin{equation}
			\sum_{i=0}^{2^{n}}\sum_{\substack{j=0\\[1mm]j\neq i}}^{2^{n}}\alpha_{i}\beta_{i}\alpha_{j}\beta_{j}\leq\sum_{i=0}^{2^{n}}\sum_{\substack{j=0\\[1mm]j\neq i}}^{2^{n}}\alpha_{i}^{2}\beta_{j}^{2}
		\end{equation}
		which, in turn, is equivalent to:
		\begin{equation}
			\begin{split}
				0 &\leq \sum_{i=0}^{2^{n}}\sum_{\substack{j=0\\[1mm]j\neq i}}^{2^{n}}\alpha_{i}^{2}\beta_{j}^{2} - \sum_{i=0}^{2^{n}}\sum_{\substack{j=0\\[1mm]j\neq i}}^{2^{n}}\alpha_{i}\beta_{i}\alpha_{j}\beta_{j}\\
				&= \dfrac{1}{2}\sum_{i=0}^{2^{n}}\sum_{\substack{j=0\\[1mm]j\neq i}}^{2^{n}}\alpha_{i}^{2}\beta_{j}^{2} + \dfrac{1}{2}\sum_{i=0}^{2^{n}}\sum_{\substack{j=0\\[1mm]j\neq i}}^{2^{n}}\alpha_{i}^{2}\beta_{j}^{2} - \sum_{i=0}^{2^{n}}\sum_{\substack{j=0\\[1mm]j\neq i}}^{2^{n}}\alpha_{i}\beta_{i}\alpha_{j}\beta_{j}\\
				&= \dfrac{1}{2}\sum_{i=0}^{2^{n}}\sum_{\substack{j=0\\[1mm]j\neq i}}^{2^{n}}(\alpha_{i}\beta_{j} - \alpha_{j}\beta_{i})^{2}.
			\end{split}
		\end{equation}
		Since this last inequality is always true, the triangle inequality is also true.
	\end{itemize}
\end{proof}
Now, a distance function $D$ can be defined for rotors.
\begin{theo}
	The function $D:\mathfrak{R}\times\mathfrak{R}\to\F{R}^{+}$ defined by the identity $D(R_{1},R_{2}) = \|R_{1}-R_{2}\|$ is a distance in $\mathfrak{R}$, i.e.,:
	\begin{itemize}
		\item[(i)] $D(R_{1},R_{2})\geq 0$ for all $R_{1},R_{2}\in\mathfrak{R}$. In particular, $D(R_{1},R_{2}) = 0$ if, and only if, $R_{1} = R_{2}$.
		\item[(ii)] $D(R_{1},R_{2}) =D(R_{2},R_{1})$ for all $R_{1},R_{2}\in\mathfrak{R}$.
		\item[(iii)] $D(R_{1},R_{3})\leq D(R_{1},R_{2}) + D(R_{2},R_{3})$ for all $R_{1},R_{2},R_{3}\in\mathfrak{R}$.
	\end{itemize}
\end{theo}
\begin{proof} \
	The proof is straightforward and uses the fact that $\|\cdot\|$ is a norm. Given two different rotors $R_{1}$ and $R_{2}$:
	\begin{itemize}
		\item[(i)] $D(R_{1},R_{2}) = \|R_{1}-R_{2}\| \geq 0$. In particular:
		\begin{equation}
			D(R_{1},R_{2}) = 0 \Longleftrightarrow \|R_{1}-R_{2}\| = 0 \stackrel{(1)}{\Longleftrightarrow} R_{1} - R_{2} = 0 \Longleftrightarrow R_{1} = R_{2},
		\end{equation}
		where $(1)$ uses the first property of a norm.
		\item[(ii)] We have that:
		\begin{equation}
			\begin{split}
				D(R_{1},R_{2}) &= \|R_{1}-R_{2}\| = \sqrt{\left<\left(R_{1}-R_{2}\right)\left(R_{1}-R_{2}\right)^\sim\right>_{0}}\\
				&= \sqrt{\left<\left(R_{2}-R_{1}\right)\left(R_{2}-R_{1}\right)^{\sim}\right>_{0}} = \|R_{2}-R_{1}\| = D(R_{2},R_{1}).
			\end{split}
		\end{equation}
		\item[(iii)] Given a third rotor $R_{3}$, we have that:
		\begin{equation}
			\begin{split}
				D(R_{1},R_{3}) &= \|R_{1}-R_{3}\| = \|R_{1}-R_{2} + R_{2} - R_{3}\|\\
				&\stackrel{(1)}{\leq}\|R_{1}-R_{2}\| + \|R_{2}-R_{3}\| = D(R_{1},R_{2}) + D(R_{2},R_{3}),
			\end{split}
		\end{equation}
		where $(1)$ uses the third property of a norm.
	\end{itemize}
\end{proof}
As stated before, the end-effector pose of a serial robot and the pose of each one of its joints are described by the configuration-dependent rotors $R(\bm{q})$ and $R_{i}(\bm{q})$ respectively. Thus, one can be tempted to extend the distance function $D$ to $\mathcal{C}$ as follows:
\begin{equation}\label{distancefirstapproach}
	\begin{split}
		&D:\mathcal{C}\times\mathcal{C}\to\F{R}^{+}\\
		&D(\bm{q}_{1},\bm{q}_{2}) = \|R(\bm{q}_{1}) - R(\bm{q}_{2})\|
	\end{split} 
\end{equation}
This function verifies all the requirements of a distance function with the exception of: 
\begin{equation}
	D(\bm{q}_{1},\bm{q}_{2}) = 0 \Longleftrightarrow \bm{q}_{1} = \bm{q}_{2}.
\end{equation}
The reason is simple: a given pose of the end-effector can have associated up to 16 different configurations if the serial robot is non-redundant and an infinite number if it is redundant. In particular, this means that $R(\bm{q}_{1}) = R(\bm{q}_{2})$ with $\bm{q}_1\neq\bm{q}_2$. However, this problem can be overcome as follows:
\begin{itemize}
	\item For each joint $i$, denote by $\mathcal{C}_{i}$ the configuration space of the subchain formed by the first $i$ joints. It is clear that, if the robot has $n$ degrees of freedom, $\mathcal{C}_{i}\subset\mathcal{C}$ for every $1\leq i\leq n$. Then, the following set of functions can be defined:
	\begin{equation}\label{distancejoint}
		\begin{split}
			D_{i}:\mathcal{C}_{i}\times\mathcal{C}_{i}&\to\F{R}^{+}\\
			D_{i}(\bm{q}_{1},\bm{q}_{2}) &= \|R_{i}(\bm{q}_{1}) - R_{i}(\bm{q}_{2})\|
		\end{split} 
	\end{equation}
	where, as stated before, $R_{i}$ is the rotor that describes the pose of joint $i$. Again, these functions are not distance functions for the same reason as $D$ (equation (\ref{distancefirstapproach})) is not a distance function.
	\item The function:
	\begin{equation}\label{distancefinal}
		\begin{split}
			&D:\mathcal{C}\times\mathcal{C}\to[0,+\infty)\\
			&D(\bm{q}_{1},\bm{q}_{2}) = D_{1}(\bm{q}_{1_{1}}\bm{q}_{2_{1}}) + \dots + D_{n}(\bm{q}_{1_{n}},\bm{q}_{2_{n}})
		\end{split} 
	\end{equation}
	where $\bm{q}_{1_{i}}$ ($\bm{q}_{2_{i}}$) denotes the first $i$ coordinates of the configuration vector $\bm{q}_{1}$ ($\bm{q}_{2}$), defines a distance function in $\mathcal{C}$.
	\begin{proof}
		Since, for each $1\leq i\leq n$, $D_{i}$ satisfies the requirements $(ii)$ and $(iii)$ of a distance function, it is clear that $D$ also satisfies them. In addition, $D_{i}(\bm{q}_{1_{i}},\bm{q}_{2_{i}})\geq 0$ for each $1\leq i\leq n$ and $\bm{q}_{1_{i}},\bm{q}_{2_{i}}\in\mathcal{C}_{i}$. Therefore, $D(\bm{q}_{1},\bm{q}_{2})\geq 0$ for arbitrary $\bm{q}_{1},\bm{q}_{2}\in\mathcal{C}$. Finally, if $D(\bm{q}_{1},\bm{q}_{2}) = 0$, then, since any term of equation (\ref{distancefinal}) is a positive scalar, it can be deduced that $D_{i}(\bm{q_{1_{i}}},\bm{q}_{2_{i}}) = 0$ for every $1\leq i\leq n$. Thus, $\bm{q}_{1}$ and $\bm{q}_{2}$ has, not only the same end-effector pose, but the pose of each of its joints, which clearly implies that $\bm{q}_{1} = \bm{q}_{2}$.
	\end{proof}
\end{itemize}

This distance function can be restricted to $\mathcal{S}$ just by considering the joints involved in a given singularity $\bm{q}_{s}$.
\begin{defi}\label{distancesingularity}
	Let $\bm{q}_{s}\in\mathcal{S}$  be a singularity of a serial robot that involves joints $i_{1},\dots,i_{r}$. Then, the function $D:\mathcal{C}\times\mathcal{S}\to\F{R}^{+}$ defined by the expression:
	\begin{equation}
		D(\bm{q},\bm{q}_{s}) = D_{i_{1}}(\bm{q}_{i_{1}},\bm{q}_{s_{i_{1}}}) + \dots + D_{i_{r}}(\bm{q}_{i_{r}},\bm{q}_{s_{i_{r}}}),
	\end{equation}
	where, for each $i_{1}\leq k\leq i_{r}$, $D_{k}$ is the function defined in (\ref{distancejoint}), is a distance function in $\mathcal{C}$.
\end{defi}

\section{Application to the serial robot Kuka LWR 4+}\label{ApplicationKuka}
\begin{figure}[t!]
	\centering
	\includegraphics[width=10cm]{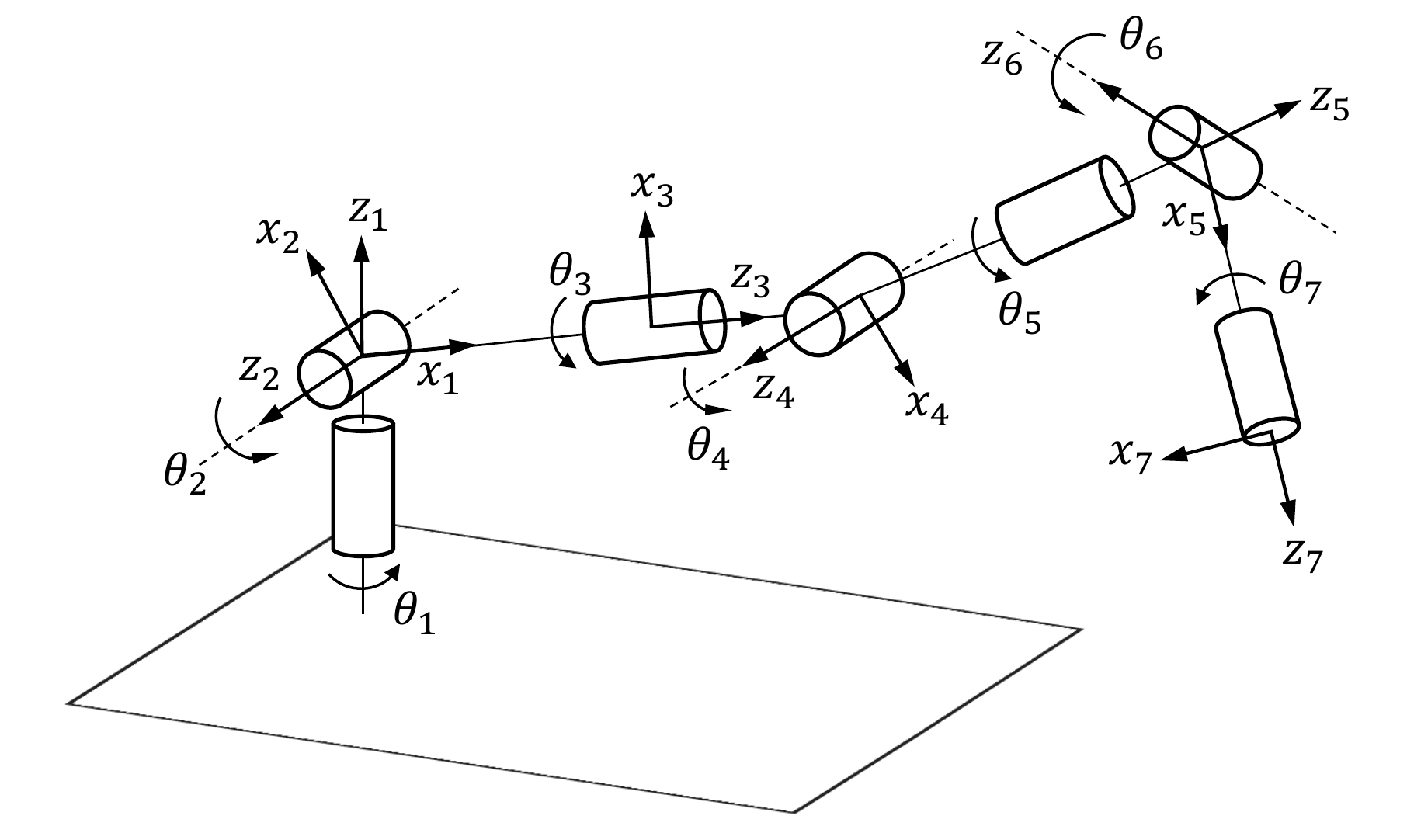}
	\caption{Schematic representation of the Kuka LWR 4+.}
	\label{Kuka_representation}
\end{figure}

To show the advantages of the proposed method, an illustrative example is developed in this section, making use of the Kuka LWR 4+, an anthropomorphic robotic arm with seven degrees of freedom and a spherical wrist. It is schematically depicted in figure \ref{Kuka_representation}. Since it has a spherical wrist, its singularities can be decoupled into position and orientation singularities. Hence, theorem \ref{theoremsimplifiedsingularities} can be applied in order to find out such singularities. The computations with the vectors of $\mathcal{G}_{3}$ have been carried out using the \textit{Clifford Multivector Toolbox} of MATLAB \cite{Sangwine17}.

With respect to the position singularities, the following system of $C(4,3) = 4$ equations should be solved:
\begin{equation}\label{eq1}
	\left.\begin{split}
		(\bm{z}_{1}\times(\bm{o}_{7}-\bm{o}_{1}))\wedge(\bm{z}_{2}\times(\bm{o}_{7}-\bm{o}_{2}))\wedge(\bm{z}_{3}\times(\bm{o}_{7}-\bm{o}_{3})) &= 0\\
		(\bm{z}_{1}\times(\bm{o}_{7}-\bm{o}_{1}))\wedge(\bm{z}_{2}\times(\bm{o}_{7}-\bm{o}_{2}))\wedge(\bm{z}_{4}\times(\bm{o}_{7}-\bm{o}_{4})) &= 0\\
		(\bm{z}_{1}\times(\bm{o}_{7}-\bm{o}_{1}))\wedge(\bm{z}_{3}\times(\bm{o}_{7}-\bm{o}_{3}))\wedge(\bm{z}_{4}\times(\bm{o}_{7}-\bm{o}_{4})) &= 0\\
		(\bm{z}_{2}\times(\bm{o}_{7}-\bm{o}_{2}))\wedge(\bm{z}_{3}\times(\bm{o}_{7}-\bm{o}_{3}))\wedge(\bm{z}_{4}\times(\bm{o}_{7}-\bm{o}_{4})) &= 0
	\end{split}\;\right\}
\end{equation}
where, as computed in \cite{ZaplanaClaretBasanez17}, we have that:
\begin{equation}
	\begin{split}
		&\bm{z}_{1}\times(\bm{o}_{7}-\bm{o}_1)\\
		&= \begin{bmatrix}
			- 400\text{c}_{2}\text{s}_{1} - 390\text{s}_{4}(\text{c}_{1}\text{s}_{3} + \text{c}_{3}\text{s}_{1}\text{s}_{2}) - 390\text{c}_{2}\text{c}_{4}\text{s}_{1}\\
			400\text{c}_{1}\text{c}_{2} - 390\text{s}_{4}(\text{s}_{1}\text{s}_{3} - \text{c}_{1}\text{c}_{3}\text{s}_{2}) + 390\text{c}_{1}\text{c}_{2}\text{c}_{4}\\
			0
		\end{bmatrix},\\
		\empty\\
		&\bm{z}_{2}\times(\bm{o}_{7}-\bm{o}_2)\\
		&= \begin{bmatrix}
			-\text{c}_{1}(400\text{s}_{2} + 390\text{c}_{4}\text{s}_{2} - 390\text{c}_{2}\text{c}_{3}\text{s}_{4})\\
			-\text{s}_{1}(400\text{s}_{2} + 390\text{c}_{4}\text{s}_{2} - 390\text{c}_{2}\text{c}_{3}\text{s}_{4})\\
			\text{c}_{1}(400\text{c}_{1}\text{c}_{2} - 390\text{s}_{4}(\text{s}_{1}\text{s}_{3} - \text{c}_{1}\text{c}_{3}\text{s}_{2}) + 390\text{c}_{1}\text{c}_{2}\text{c}_{4}) + \text{s}_{1}(400\text{c}_{2}\text{s}_{1} + 390\text{s}_{4}(\text{c}_{1}\text{s}_{3} + \text{c}_{3}\text{s}_{1}\text{s}_{2}) + 390\text{c}_{2}\text{c}_{4}\text{s}_{1})
		\end{bmatrix},\\
		\empty\\
		&\bm{z}_{3}\times(\bm{o}_{7}-\bm{o}_3)\\
		&= \begin{bmatrix}
			\text{c}_{2}\text{s}_{1}(390\text{c}_{4}\text{s}_{2} - 390\text{c}_{2}\text{c}_{3}\text{s}_{4}) - \text{s}_{2}(390\text{s}_{4}(\text{c}_{1}\text{s}_{3} + \text{c}_{3}\text{s}_{1}\text{s}_{2}) + 390\text{c}_{2}\text{c}_{4}\text{s}_{1})\\
			- \text{s}_{2}(390\text{s}_{4}(\text{s}_{1}\text{s}_{3} - \text{c}_{1}\text{c}_{3}\text{s}_{2}) - 390\text{c}_{1}\text{c}_{2}\text{c}_{4}) - \text{c}_{1}\text{c}_{2}(390\text{c}_{4}\text{s}_{2} - 390\text{c}_{2}\text{c}_{3}\text{s}_{4})\\
			\text{c}_{1}\text{c}_{2}(390\text{s}_{4}(\text{c}_{1}\text{s}_{3} + \text{c}_{3}\text{s}_{1}\text{s}_{2}) + 390\text{c}_{2}\text{c}_{4}\text{s}_{1}) + \text{c}_{2}\text{s}_{1}(390\text{s}_{4}(\text{s}_{1}\text{s}_{3} - \text{c}_{1}\text{c}_{3}\text{s}_{2}) - 390\text{c}_{1}\text{c}_{2}\text{c}_{4})
		\end{bmatrix},\\
		\empty\\
		&\bm{z}_{4}\times(\bm{o}_{7}-\bm{o}_4)\\
		&= \begin{bmatrix}
			(390\text{c}_{4}\text{s}_{2} - 390\text{c}_{2}\text{c}_{3}\text{s}_{4})(\text{c}_{1}\text{c}_{3} - \text{s}_{1}\text{s}_{2}\text{s}_{3}) - \text{c}_{2}\text{s}_{3}(390\text{s}_{4}(\text{c}_{1}\text{s}_{3} + \text{c}_{3}\text{s}_{1}\text{s}_{2}) + 390\text{c}_{2}\text{c}_{4}\text{s}_{1})\\
			(390\text{c}_{4}\text{s}_{2} - 390\text{c}_{2}\text{c}_{3}\text{s}_{4})(\text{c}_{3}\text{s}_{1} + \text{c}_{1}\text{s}_{2}\text{s}_{3}) - \text{c}_{2}\text{s}_{3}(390\text{s}_{4}(\text{s}_{1}\text{s}_{3} - \text{c}_{1}\text{c}_{3}\text{s}_{2}) - 390\text{c}_{1}\text{c}_{2}\text{c}_{4})\\
			(\text{c}_{1}\text{c}_{3} - \text{s}_{1}\text{s}_{2}\text{s}_{3})(390\text{s}_{4}(\text{s}_{1}\text{s}_{3} - \text{c}_{1}\text{c}_{3}\text{s}_{2}) - 390\text{c}_{1}\text{c}_{2}\text{c}_{4}) - (390\text{s}_{4}(\text{c}_{1}\text{s}_{3} + \text{c}_{3}\text{s}_{1}\text{s}_{2}) + 390\text{c}_{2}\text{c}_{4}\text{s}_{1})(\text{c}_{3}\text{s}_{1} + \text{c}_{1}\text{s}_{2}\text{s}_{3})
		\end{bmatrix},
	\end{split}
\end{equation}
with $\text{c}_i = \cos(\theta_i)$ and $\text{s}_i = \sin(\theta_i)$. However, in order to simplify these expressions, the system of equations (\ref{eq1}) is expressed with respect to the frame attached to the fourth joint of the Kuka LWR 4+. To do so, a relation analogous of relation (\ref{B}) is applied. Here, instead of pre-multiplying by the corresponding rotation matrix, the system of equations (\ref{eq1}) is multiplied by the three-dimensional rotor $R$ that performs the rotation between the frame attached to the end-effector and the frame attached to the fourth joint. For instance, the first equation of the system (\ref{eq1}) becomes:
\begin{equation}
	R(\bm{z}_{1}\times(\bm{o}_{7}-\bm{o}_{1}))\wedge(\bm{z}_{2}\times(\bm{o}_{7}-\bm{o}_{2}))\wedge(\bm{z}_{3}\times(\bm{o}_{7}-\bm{o}_{3}))\widetilde{R} = 0,
\end{equation}
which, using the geometric covariance property for rotors introduced in section \ref{Background}, becomes:
\begin{equation}
	R(\bm{z}_{1}\times(\bm{o}_{7}-\bm{o}_{1}))\widetilde{R}\wedge R(\bm{z}_{2}\times(\bm{o}_{7}-\bm{o}_{2}))\widetilde{R}\wedge R(\bm{z}_{3}\times(\bm{o}_{7}-\bm{o}_{3}))\widetilde{R} = 0.
\end{equation}
Therefore, the system of equations (\ref{eq1}) becomes:
\begin{equation}\label{eq1bis}
	\left.\begin{split}
		a_{1}\wedge a_{2}\wedge a_{3} &= 0\\
		a_{1}\wedge a_{2}\wedge a_{4} &= 0\\
		a_{1}\wedge a_{3}\wedge a_{4} &= 0\\
		a_{2}\wedge a_{3}\wedge a_{4} &= 0
	\end{split}\;\right\}
\end{equation}
where
\begin{equation}
	\begin{split}
		a_{1} &= (-10\text{c}_{2}\text{s}_{3}(40\text{c}_{4} + 39))e_{1}+(400\text{c}_{2}\text{s}_{3}\text{s}_{4})e_{2}+(400\text{c}_{2}\text{c}_{3} + 390\text{s}_{2}\text{s}_{4} + 390\text{c}_{2}\text{c}_{3}\text{c}_{4})e_{3},\\
		a_{2} &= (10\text{c}_{3}(40\text{c}_{4} + 39))e_{1}+ (-400\text{c}_{3}\text{s}_{4})e_{2}+(10\text{s}_{3}(40\text{c}_{4} + 39))e_{3},\\
		a_{3} &= (390\text{s}_{4})e_{3},\\
		a_{4} &= (-390)e_{1}.
	\end{split}
\end{equation}
Now, the system of equations (\ref{eq1bis}) becomes:
\begin{equation}
	\left.\begin{split}
		0 &= 0\\
		40\text{c}_{2}\text{s}_{4} + 39\text{s}_{2}\text{c}_{3}\text{s}_{4}^{2}+39\text{c}_{2}\text{c}_{4}\text{s}_{4} &= 0\quad\\
		\text{c}_{2}\text{s}_{3}\text{s}_{4}^{2} &= 0\\
		\text{c}_{3}\text{s}_{4}^{2} &= 0
	\end{split}\right\}
\end{equation}
which clearly has two different solutions:
\begin{itemize}
	\item $\text{s}_{4} = 0$ or, equivalently, $q_{4} = 0$.
	\item $\text{c}_{2} = \text{c}_{3} = 0$ or, equivalently, $q_{2} =\pm\frac{\pi}{2} $ and $q_{3} = \pm\frac{\pi}{2}$.
\end{itemize}
These two solutions correspond to the position singularities of the Kuka LWR 4+.

With respect to the orientation singularities, there is only one equation to solve:
\begin{equation}\label{eqOrSing}
	\bm{z}_{5}\wedge\bm{z}_{6}\wedge\bm{z}_{7} = 0.
\end{equation}
Again, the expression of each $\bm{z}_{i}$ for $i=5,6,7$ can be simplified by expressing those vectors with respect to the frame attached to the fourth joint. Thus, equation (\ref{eqOrSing}) becomes:
\begin{equation}\label{expandedsingularity}
	\begin{split}
		&e_{2}\wedge(-\text{s}_{5}e_{1}-\text{c}_{5}e_{3})\wedge(\text{c}_{5}\text{s}_{6}e_{1}+\text{c}_{6}e_{2}-\text{s}_{5}\text{s}_{6}e_{3})\\
		&= (-\text{s}_{5}e_{2}\wedge e_{1}-\text{c}_{5}e_{2}\wedge e_{3})\wedge(\text{c}_{5}\text{s}_{6}e_{1}+\text{c}_{6}e_{2}-\text{s}_{5}\text{s}_{6}e_{3})\\
		&\stackrel{(1)}{=} -\text{s}_{5}^{2}\text{s}_{6}e_{1}\wedge e_{2}\wedge e_{3} - \text{c}_{5}^{2}\text{s}_{6}e_{1}\wedge e_{2}\wedge e_{3} = -\text{s}_{6}e_{1}\wedge e_{2}\wedge e_{3} = 0,
	\end{split}
\end{equation}
where $(1)$ uses the anticommutativity of the outer product. Clearly, the last expression of equation (\ref{expandedsingularity}) is zero if, and only if, $\text{s}_{6} = 0$ or, equivalently, if, and only if, $q_{6}=0$. Thus, the Kuka LWR 4+ only has one orientation singularity (the wrist singularity, as explained in remark \ref{remwristsingularity}).

Finally, the distance function defined in \ref{distancesingularity} can be applied to any of the already obtained singular configurations. Let us consider, for instance, the position singularity $q_{4}=0$. Then, the distance between an arbitrary configuration $\bm{q}\in\mathcal{C}$ and this singularity is given by the expression:
\begin{equation}\label{distanceq4}
	D(\bm{q},\bm{q}_{s}) = \|R_{4}(\bm{q})-R_{4}(\bm{q}_{s})\|,
\end{equation}
where $\bm{q}_{s}$ denotes the singular configuration $q_{4}=0$ and $R_{4}$ is the rotor defining the pose of the fourth joint of the Kuka LWR 4+. 

In particular, $R_{4}$ can be found as explained in section \ref{Distance}. Indeed, if $\{e_{1},e_{2},e_{3}\}$ denotes the orthogonal basis defined by the world frame and $\{f_1,f_2,f_3\}$ (resp. $\{f'_1, f'_2,f'_3\}$), the orthogonal basis defined by the frame attached to the fourth joint under the effect of configuration $\bm{q}$ (resp. singular configuration $\bm{q}_s$), then:
\begin{equation}\label{Rotor4Chp6}
	\begin{split}
		R_{4}(\bm{q}) &= \dfrac{1 + e^{1}f_{1}+e^{2}f_{2}+e^{3}f_{3}}{\|1 + e^{1}f_{1}+e^{2}f_{2}+e^{3}f_{3}\|},\\
		\empty\\
		R_{4}(\bm{q}_s) &= \dfrac{1 + e^{1}f'_{1}+e^{2}f'_{2}+e^{3}f'_{3}}{\|1 + e^{1}f'_{1}+e^{2}f'_{2}+e^{3}f'_{3}\|},
	\end{split}
\end{equation}
where $\{e^{1},e^{2},e^{3}\}$ is the reciprocal frame \cite{DoranLasenby03,LavorXamboZaplana18} of $\{e_{1},e_{2},e_{3}\}$. Since $\{e_{1},e_{2},e_{3}\}$ is also an orthonormal set of vectors, such a reciprocal frame is:
\begin{equation}
	\begin{split}
		e^{1} &= e_{1},\\
		e^{2} &= e_{2},\\
		e^{3} &= e_{3}.
	\end{split}
\end{equation}
Thus, equation (\ref{Rotor4Chp6}) turns to:
\begin{equation}\label{Rotor4Chp6Bis}
	\begin{split}
		R_{4}(\bm{q}) &= \dfrac{1 + e_{1}f_{1}+e_{2}f_{2}+e_{3}f_{3}}{\|1 + e_{1}f_{1}+e_{2}f_{2}+e_{3}f_{3}\|},\\
		\empty\\
		R_{4}(\bm{q}_s) &= \dfrac{1 + e_{1}f'_{1}+e_{2}f'_{2}+e_{3}f'_{3}}{\|1 + e_{1}f'_{1}+e_{2}f'_{2}+e_{3}f'_{3}\|}.
	\end{split}
\end{equation}
Evaluating equation (\ref{Rotor4Chp6Bis}) for the KUKA LWR 4+, we obtain:
\begin{equation}
	\begin{split}
		R_{4}(\bm{q}) &=\dfrac{a_{1}+a_{2}e_{1}\wedge e_{2}+a_{3}e_{1}\wedge e_{3}+a_{4}e_{2}\wedge e_{3}}{\sqrt{a_1^2+a_2^2+a_3^2+a_4^2}},\\
		\empty\\
		R_{4}(\bm{q}_{s}) &=\dfrac{b_{1}+b_{2}e_{1}\wedge e_{2}+b_{3}e_{1}\wedge e_{3}+b_{4}e_{2}\wedge e_{3}}{\sqrt{b_1^2+b_2^2+b_3^2+b_4^2}},
	\end{split}
\end{equation}
where $\{e_{1}\wedge e_{2},e_{1}\wedge e_{3},e_{2}\wedge e_{3}\}$ are the basis bivectors of $\mathcal{G}_{3}$ and
\begin{equation}
	\begin{split}
		a_{1} &=\text{c}_{2}\text{s}_{3} + \text{c}_{4}\text{s}_{1}\text{s}_{3}-\text{c}_{4}\text{c}_{3}\text{c}_{1}\text{s}_{2}+\text{s}_{3}\text{s}_{4}\text{c}_{1} + \text{s}_{4}\text{s}_{1}\text{s}_{2}\text{c}_{3}+\text{s}_{4}\text{c}_{1}\text{c}_{2}+\text{c}_{2}\text{c}_{4}\text{s}_{1},\\
		a_{2} &=\text{c}_{2}\text{s}_{1}\text{s}_{4}-\text{c}_{4}\text{c}_{1}\text{s}_{3}-\text{c}_{4}\text{c}_{3}\text{s}_{1}\text{s}_{2}-\text{c}_{1}\text{c}_{2}\text{c}_{4}+\text{s}_{4}\text{s}_{1}\text{s}_{3}-\text{s}_{4}\text{s}_{2}\text{c}_{1}\text{c}_{3},\\
		a_{3} &=\text{s}_{2}\text{s}_{4}+\text{c}_{2}\text{c}_{3}\text{c}_{4}+\text{c}_{3}\text{s}_{1}+\text{c}_{1}\text{s}_{2}\text{s}_{3},\\
		a_{4} &=\text{c}_{4}\text{s}_{2}-\text{c}_{2}\text{c}_{3}\text{s}_{4}-\text{c}_{1}\text{c}_{3}+\text{s}_{1}\text{s}_{2}\text{s}_{3},\\
		b_{1} &=\text{c}_{2}\text{s}_{3} + \text{s}_{1}\text{s}_{3}-\text{c}_{3}\text{c}_{1}\text{s}_{2}+\text{c}_{2}\text{s}_{1},\\
		b_{2} &=-\text{c}_{1}\text{s}_{3}-\text{c}_{3}\text{s}_{1}\text{s}_{2}-\text{c}_{1}\text{c}_{2},\\
		b_{3} &= \text{c}_{2}\text{c}_{3}+\text{c}_{3}\text{s}_{1}+\text{c}_{1}\text{s}_{2}\text{s}_{3},\\
		b_{4} &=\text{s}_{2}-\text{c}_{1}\text{c}_{3}+\text{s}_{1}\text{s}_{2}\text{s}_{3}.
	\end{split}
\end{equation}
Therefore, by proposition \ref{propositionnorma} and the decomposition used in the proof of lemma \ref{lemmanorma}, the distance of an arbitrary configuration $\bm{q}$ to the position singularity $q_{4}=0$ is given by:
\begin{equation}
	D(\bm{q},\bm{q}_{s})= \sqrt{(a'_{1}-b'_{1})^{2}+(a'_{2}-b'_{2})^{2}+(a'_{3}-b'_{3})^{2}+(a'_{4}-b'_{4})^{2}},
\end{equation}
where
\begin{equation}
	a'_i = \dfrac{a_i}{\sqrt{a_1^2+a_2^2+a_3^2+a_4^2}}\text{ and }b'_i = \dfrac{b_i}{\sqrt{b_1^2+b_2^2+b_3^2+b_4^2}}.
\end{equation}

\section{Handling of singularities}\label{Handling}
Once the set of singular configurations $\mathcal{S}$ has been identify, several methods can be applied to handle the singularities. The detailed treatment of this topic is beyond the scope of this work. However, in order to show the possibilities of the distance function  proposed in section \ref{Distance}, we comment on three different situations, namely motion planning, motion control and bilateral teleoperation. In each one of these situations, the distance function defined in \ref{distancesingularity} plays an important role for handling the singularities.

\subsection{Singularity handling in motion planning}
Motion planning consists of programming collision-free motions for a given robotic manipulator from a start position to a goal position among a collection of static obstacles. The subset of robot configurations that do not cause collision with such obstacles is termed \textit{free-of-obstacles configuration space} and it is denoted by $\mathcal{C}_{\text{free}}$. The main methods used for motion planning can be grouped in three categories:
\begin{itemize}
	\item Potential field methods, where a differentiable real-valued function $U:\mathcal{C}\to\mathbb{R}$, called the potential function, is defined. Such a function has an attractive component that pulls the trajectory towards the goal configuration and a repulsive component that pushes the trajectory away from the start configuration and from the obstacles.
	\item Sampling-based multi-query methods, where a roadmap is constructed over $\mathcal{C}_{\text{free}}$. The nodes represent free-of-obstacles configurations, while the edges represent feasible local paths between those configurations. Once the roadmap is constructed, a search algorithm finds out the best solution trajectory by selecting and joining the local paths through an optimization process. 
	\item Sampling-based single-query methods, where a tree-structure data is constructed by searching new configurations (nodes) in $\mathcal{C}_{\text{free}}$ and connecting them through local paths (edges). Its main difference with respect to the multi-query methods is that, while the multi-query methods work in two steps (construction of the roadmap and searching of a solution trajectory),  in the single-query methods both steps are taken together. Each new configuration added to the set of nodes is connected by a local path and evaluated in order to check its feasibility.
\end{itemize}

For any method of these three categories, the distance function $D$ defined in \ref{distancesingularity} can be applied to construct solution trajectories that also avoid the singularities. Indeed:
\begin{itemize}
	\item For a potential field method, it is sufficient to add a repulsive component that pushes the trajectory, not only away from obstacles, but also away from singularities. To do so, the most efficient way is to define, for each singularity $\bm{q}_{s}$, a quadratic repulsive component as follows:
	\begin{equation}
		U_{r,\bm{q}_{s}}(\bm{q}) = \left\{\begin{array}{ll}
			\dfrac{\kappa}{2}\left(\dfrac{1}{D(\bm{q},\bm{q}_{s})}-\dfrac{1}{D_{0}}\right)^{2} & \text{if }D(\bm{q},\bm{q}_{s})\leq D_{0}\\
			0 & \text{if }D(\bm{q},\bm{q}_{s})>D_{0}
		\end{array}\right.
	\end{equation}
	where $D_{0}$ is set as a threshold for the distance $D$ and $\kappa\in\mathbb{R}$. 
	\item For a sampling-based method with multiple queries, it is sufficient to remove from the roadmap those nodes associated with singular configurations. During the construction of the roadmap, each configuration $\bm{q}\in\mathcal{C}$ is evaluated to determine whether $\bm{q}$ is free-of-obstacles or not. Similarly, the idea is to evaluate each $\bm{q}\in\mathcal{C}$ in order to determine whether $\bm{q}$ is close to a singularity or not. To speed up the process, both evaluations can be carried out together:
	\begin{itemize}
		\item[1)] Select a value $D_{0}>0$ that will work as a threshold.
		\item[2)] Given a discretization of the configuration space $\mathcal{C}$, each $\bm{q}$ of this discretization is evaluated to check whether:
		\begin{itemize}
			\item It is free-of-obstacles.
			\item It is far from any singularity. This can be done simply by evaluating whether $D(\bm{q},\bm{q}_{s})>D_{0}$ or $D(\bm{q},\bm{q}_{s})\leq D_{0}$.
		\end{itemize}
		\item[3)] If $\bm{q}$ is free-of-obstacles and far from any singularity, then it can be added to the set of nodes of the roadmap.
	\end{itemize}
	\item For a sampling-based method with a single query, the approach is completely analogous to the one used for methods with multiple-queries due to the similarities between both categories.
\end{itemize}

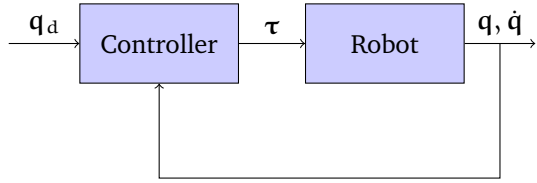
\begin{figure}[t!]
	\begin{center}
		\tikzstyle{block} = [draw, fill=blue!20, rectangle, 
		minimum height=3em, minimum width=6em]
		\tikzstyle{sum} = [draw, fill=blue!20, circle, node distance=1cm]
		\tikzstyle{input} = [coordinate]
		\tikzstyle{output} = [coordinate]
		\tikzstyle{pinstyle} = [pin edge={to-,thin,black}]
		
		% The block diagram code is probably more verbose than necessary
		\begin{tikzpicture}[auto, node distance=2cm]
			% We start by placing the blocks
			\node [input, name=input] {};
			\node [block, right of=input] (controller) {Controller};
			\node [block, right of=controller,
			node distance=3cm] (robot) {Robot};
			% We draw an edge between the controller and system block to 
			% calculate the coordinate u. We need it to place the measurement block. 
			\draw [->] (controller) -- node[name=u] {$\bm{\tau}$} (robot);
			\node [output, right of=robot] (output) {};
			\node [input, below of=u] (measurements) {};
			
			% Once the nodes are placed, connecting them is easy. 
			\draw [->] (input) -- node {$\bm{q}_{d}$} (controller);
			\draw [->] (robot) -- node [name=y] {$\bm{q},\dot{\bm{q}}$}(output);
			\draw [-] (y) |- (measurements);
			\draw [->] (measurements) -|  (controller);
		\end{tikzpicture}
	\end{center}
	\caption{Standard motion control scheme.}
	\label{control}
\end{figure}

\subsection{Singularity handling in motion control}
Motion control consists of making the end-effector of a robot follow a time-varying trajectory specified within the manipulator workspace. A typical Inverse Dynamics Control scheme (depicted as a block diagram in figure \ref{control}) can be described as:
\begin{itemize}
	\item An input, i.e., the desired or target configuration $\bm{q}_{d}$ together with its velocity $\dot{\bm{q}}_{d}$.
	\item A controller based on the dynamical model of the robot:
	\begin{equation}\label{dinamicalmodel}
		\bm{\tau} = M(\bm{q})\ddot{\bm{q}}+C(\bm{q},\dot{\bm{q}})\dot{\bm{q}} +g(\bm{q})
	\end{equation}
	where $M(\bm{q})$ denotes the inertia matrix of the robot, $C(\bm{q},\dot{\bm{q}})$ denotes the matrix of Coriolis and centrifugal forces and $g(\bm{q})$, the gravity vector.
	\item An output, i.e., the vector of torques $\bm{\tau}$, that is sent to the robot to perform the desired motion.
	\item The robot executes the motion and updates the vectors $\bm{q}$ and $\dot{\bm{q}}$.
	\item The robot sends such updated vectors to the controller (also known as the \textit{feedback} of the system).
\end{itemize}

\begin{figure}[t!]
	\begin{center}
		\tikzstyle{block} = [draw, fill=blue!20, rectangle, minimum height=3em, minimum width=6em]
		\tikzstyle{decision} = [diamond, minimum width=3em, minimum height=1em, text centered, draw=black, fill=blue!20,inner sep =-0.1cm,aspect=1.25,text width=3cm]
		\tikzstyle{sum} = [draw, fill=blue!20, circle, node distance=1cm]
		\tikzstyle{input} = [coordinate]
		\tikzstyle{output} = [coordinate]
		\tikzstyle{pinstyle} = [pin edge={to-,thin,black}]
		
		% The block diagram code is probably more verbose than necessary
		\begin{tikzpicture}[auto, node distance=2cm]
			% We start by placing the blocks
			\node [input, name=input] {};
			\node [decision,right of=input] (sing) {$D(\bm{q}_{d},\bm{q}_{s})>D_{0}$};
			\node [block, right of=sing,node distance=4cm] (DM) {Dynamical model};
			% We draw an edge between the controller and system block to 
			% calculate the coordinate u. We need it to place the measurement block. 
			\node [output, right of=DM,node distance=2.5cm] (output) {};
			\node [block, below of=sing] (measurements) {$\overline{\bm{q}}_{d}=\bm{q}_{d}+D_{0}\bm{q}_{d}$};
			
			% Once the nodes are placed, connecting them is easy. 
			\draw [->] (input) -- node {$\bm{q}_{d}$} (sing);
			\draw [->] (DM) -- node [name=y] {$\bm{\tau}$}(output);
			\draw [->] (sing) -- node[anchor=north] {yes} (DM);
			\draw [->] 	(sing) -- node[anchor=east] {no} (measurements);
			\draw [-] (measurements) -|  (input);
		\end{tikzpicture}
	\end{center}
	\caption{Proposed control scheme in presence of singularities.}
	\label{controlsing}
\end{figure}
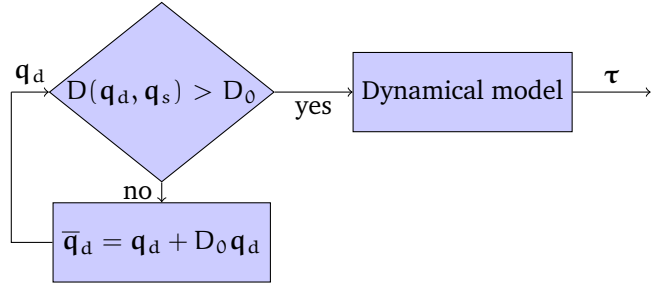

To handle the singularities, a restriction can be defined inside the controller:
\begin{itemize}
	\item[1)] The target configuration $\bm{q}_{d}$ enters in the controller.
	\item[2)] $\bm{q}_{d}$ is checked in order to determine whether it is close to a singularity or not:
	\begin{itemize}
		\item Select a threshold value $D_{0}>0$.
		\item Evaluate the condition $D(\bm{q}_{d},\bm{q}_{s})>D_{0}$ for each singularity $\bm{q}_{s}$.
		\item If the evaluation returns yes, then $\bm{\tau}$ can be computed from $\bm{q}_{d}$ using the dynamical model (equation (\ref{dinamicalmodel})) and sent to the robot. Otherwise, $\bm{q}_{d}$ is substituted by $\bm{q}_{d}+D_{0}\bm{q}_{d}$ and evaluated again.
	\end{itemize}
\end{itemize}
A block diagram of this scheme is depicted in figure \ref{controlsing}.

\subsection{Singularity handling in bilateral teleoperation}
Teleoperated robotic systems are characterized by a robot that executes the movements/actions commanded by a human operator. Any high-level or planning decision is made by a human user, while the robot is responsible for their mechanical implementation \cite{Basanez09}. Teleoperation systems are often, at least conceptually, split into two parts: a local manipulator and a remote manipulator. The first one refers to the device moved by the human operator, while the second refers to the robot or robot system that performs the action.

According to the information flow direction, the teleoperation may be unilateral or bilateral. In unilateral teleoperation, the local manipulator sends position or force data to the remote manipulator and only receives, as feedback, visual information from the remote scene. But, in bilateral teleoperation, position or force data are also sent from the remote manipulator in addition to the visual information. 

In a bilateral teleoperation system, some strategies for handling of kinematic singularities can make use of the distance function $D$ defined in \ref{distancesingularity}. For instance, the following scheme could be applied:
\begin{itemize}
	\item[1)] Select a value $D_{0}>0$ that will work as a threshold.
	\item[2)] The local manipulator sends a pose (force) $\bm{p}$ ($\bm{f}$) to the remote manipulator.
	\item[3)] The controller of the remote manipulator obtains the associated configuration $\bm{q}$.
	\item[4)] The distance $D(\bm{q},\bm{q}_{s})$ is computed for each $\bm{q}_{s}\in\mathcal{S}$.
	\item[5)] If, for some $\bm{q}_{s}$, $D(\bm{q},\bm{q}_{s})<D_{0}$, then the remote manipulator controller computes a reaction force $\bm{f}_{s}$ in the same direction of the motion but with inverse sense.
	\item[6)] The remote manipulator sends this force $\bm{f}_{s}$ to the local manipulator.
	\item[7)] The human operator will not be able to move the local manipulator in such a direction implied by $\bm{f}_{s}$ and, thus, the singularity $\bm{q}_{s}$ will never be reached.
\end{itemize}

\section{Conclusions}\label{Conclusions}
This paper proposes a novel singularity identification method for arbitrary serial robots based on the six-dimensional geometric algebra $\mathcal{G}_6$. For non-redundant serial robots, we take the six unit twists $\xi_1,\dots,\xi_6$ associated with the joints and we model them as vectors of $\mathcal{G}_6$. Hence, the problem reduces to find the configurations causing the exterior product $\xi_1(\bm{q})\wedge\cdots\wedge\xi_6(\bm{q})$ to vanish since, as proven in corollary \ref{caracterizacionsingularidades}, $\xi_1(\bm{q})\wedge\cdots\wedge\xi_6(\bm{q}) = 0$ if, and only if, $\bm{q}\in\mathcal{S}$ . Analogously, for a redundant robot with $n$ DoF, we consider the $C(n,6)$ different combinations of six unit twists taken from $\{\xi_1,\dots,\xi_n\}$ and we find the configurations causing all the exterior products of the form $\xi_{j_1}(\bm{q})\wedge\cdots\wedge\xi_{j_6}(\bm{q})$ for $1\leq j\leq C(n,6)$ to vanish.

For serial robots with a spherical wrist, a simplification is possible. For these manipulators, the singularities are of two types: position singularities and orientation singularities. The former are identified as the configurations causing the exterior products $\bm{s}_{i_{1}}(\bm{q})\wedge \bm{s}_{i_{2}}(\bm{q})\wedge \bm{s}_{i_{3}}(\bm{q})$ to vanish for $1\leq i\leq C(n-3,3)$, where $\bm{s}_{i_{j}}$ is the linear velocity component of the unit twist $\xi_{i_{j}}$ and is modelled as a vector of $\mathcal{G}_3$, while the latter are identified as the configuration causing the exterior product $\bm{z}_{n-2}(\bm{q})\wedge\bm{z}_{n-1}(\bm{q})\wedge\bm{z}_{n}(\bm{q})$ to vanish, where $\bm{z}_{i}$ is the $i$-th joint axis and, again, is modelled as a vector of $\mathcal{G}_3$. Thus, the simplification consists of evaluating the exterior product of three vectors in $\mathcal{G}_3$, instead of six vectors in $\mathcal{G}_6$.

Once the singularities are identified, a distance function is defined such as its restriction to the singular set $\mathcal{S}$, defined in \ref{distancesingularity}, is also a distance function that allows us to check how far an arbitrary configuration $\bm{q}$ is to a singularity. This distance function exploits the fact that between any two vectors $\bm{x},\bm{y}\in\mathcal{G}_n$, there always exists a rotor $R$ such that $\bm{y} = R\bm{x}\widetilde{R}$.

The advantages of the strategy introduced in this work are clear. First, it is a computer-friendly approach that avoids the computation of the determinant of an order $6\times n$ (for non-redundant robots) or $n\times n$ (for redundant robots) matrix, the Jacobian matrix $J$. In addition, the novel distance function defined in \ref{distancesingularity} can be used to improve the performance of current control schemes or motion planning algorithms which, as seen in the introduction, is still a hot research topic in robotics.

\section*{References}
\bibliography{MiBibliografia}%

\begin{thebibliography}{10}
\expandafter\ifx\csname url\endcsname\relax
  \def\url#1{\texttt{#1}}\fi
\expandafter\ifx\csname urlprefix\endcsname\relax\def\urlprefix{URL }\fi
\expandafter\ifx\csname href\endcsname\relax
  \def\href#1#2{#2} \def\path#1{#1}\fi

\bibitem{Gottlieb86}
D.~Gottlieb, Robots and topology, IEEE International Conference on Robotics and
  Automation (ICRA), San Francisco, CA, USA, April 7-10, 1986, pp. 1689--1691.

\bibitem{Hollerbach85}
J.~Hollerbach, Optimum kinematic design for a seven degree of freedom
  manipulator, in: H.~Hanafusa, H.~Inoue (Eds.), Robotics Research: The Second
  International Symposium, MIT Press, 1985, pp. 215--222.

\bibitem{Carmicheletal2020}
M.~Carmichael, R.~Khonasty, S.~Aldini, D.~Liu, Human preferences in using
  damping to manage singularities during physical human-robot collaboration,
  IEEE International Conference on Robotics and Automation (ICRA), Paris,
  France, 31 May-31 August, 2020, pp. 10184--10190.

\bibitem{Thananjeyanetal2019}
B.~Thananjeyan, A.~Tanwani, J.~Ji, D.~Fer, V.~Patel, S.~Krishnan, K.~Goldberg,
  Optimizing robot-assisted surgery suture plans to avoid joint limits and
  singularities, International Symposium on Medical Robotics (ISMR), Atlanta,
  USA, April 3-5, 2019, pp. 1--7.

\bibitem{Dupac2018}
M.~Dupac, Smooth trajectory generation for rotating extensible manipulators,
  Mathematical Methods in the Applied Sciences 41~(6) (2018) 2281--2286.

\bibitem{Wangetal2018}
X.~Wang, D.~Zhang, C.~Zhao, H.~Zhang, H.~Yan, {Singularity analysis and
  treatment for a 7R 6-DOF painting robot with non-spherical wrist}, Mechanism
  and Machine Theory 126 (2018) 92--107.

\bibitem{Ratajczak2020}
J.~Ratajczak, K.~Tcho\'{n}, Normal forms and singularities of non-holonomic
  robotic systems: A study of free-floating space robots, Systems \& Control
  Letters 138 (2020) 104661.

\bibitem{Almarkhi2019}
A.~Almarkhi, A.~Maciejewski, Singularity analysis for redundant manipulators of
  arbitrary kinematic structure, Proceedings of the 16th International
  Conference on Informatics in Control, Automation and Robotics (ICINCO),
  Prague, Czech Republic, July 29-31, 2019, pp. 42--49.

\bibitem{sharifi2019identification}
H.~Sharifi, W.~Black, Identification algorithm to determine the trajectory of
  robots with singularities, \url{http://arxiv.org/abs/1911.06632} (2019).
\newblock \href {http://arxiv.org/abs/arXiv:1911.06632}
  {\path{arXiv:arXiv:1911.06632}}.

\bibitem{Hugo2020}
H.~Hadfield, L.~Wei, J.~Lasenby, The forward and inverse kinematics of a
  {D}elta robot, in: N.~Magnenat-Thalmann, C.~Stephanidis, E.~Wu, D.~Thalmann,
  B.~Sheng, J.~Kim, G.~Papagiannakis, M.~Gavrilova (Eds.), Advances in Computer
  Graphics, Springer International Publishing, 2020, pp. 447--458.

\bibitem{Thiruvengadam2021}
S.~Thiruvengadam, J.~Tan, K.~Miller, A generalised quaternion and clifford
  algebra based mathematical methodology to effect multi-stage reassembling
  transformations in parallel robots, Advances in Applied Clifford Algebras
  31~(3) (2021) 39.

\bibitem{CorrochanoSobczyk01}
E.~Corrochano, G.~Sobczyk, {Applications of Lie algebras and the algebra of
  incidence}, in: E.~Corrochano, G.~Sobczyk (Eds.), Geometric Algebra with
  Applications in Science and Engineering, Birkh{\"a}user Boston, Boston, MA,
  2001, pp. 252--277.

\bibitem{KanaanWengerCaroChablat09}
D.~Kanaan, P.~Wenger, S.~Caro, D.~Chablat, {Singularity analysis of lower
  mobility parallel manipulators using Grassmann--Cayley algebra}, IEEE
  Transactions on Robotics 25~(5) (2009) 995--1004.

\bibitem{Tanev06}
T.~Tanev, {Singularity analysis of a 4-DOF parallel manipulator using geometric
  algebra}, in: J.~Lennar{\v{c}}i{\v{c}}, B.~Roth (Eds.), Advances in Robot
  Kinematics: Mechanisms and Motion, Springer Netherlands, Dordrecht, 2006, pp.
  275--284.

\bibitem{ChaiXiang16}
X.~Chai, J.~Xiang, Mobility analysis of limited-degrees-of-freedom parallel
  mechanisms in the framework of geometric algebra, ASME Journal of Mechanisms
  and Robotics 8~(4) (2016) 41005--41005/9.

\bibitem{YaoChenChaiLi17}
H.~Yao, Q.~Chen, X.~Chai, Q.~Li, {Singularity analysis of 3-RPR parallel
  manipulators using geometric algebra}, Advances in Applied Clifford Algebras
  27~(3) (2017) 2097--2113.

\bibitem{ChaiLi17}
X.~Chai, Q.~Li, {Analytical mobility analysis of Bennett linkage using
  geometric algebra}, Advances in Applied Clifford Algebras 27~(3) (2017)
  2083--2095.

\bibitem{Maetal2017}
J.~Ma, Q.~Chen, H.~Yao, X.~Chai, Q.~Li, Singularity analysis of the 3/6 stewart
  parallel manipulator using geometric algebra, Mathematical Methods in the
  Applied Sciences 41~(6) (2018) 2494--2506.

\bibitem{HuoSunSong17}
X.~Huo, T.~Sun, Y.~Song, A geometric algebra approach to determine
  motion/constraint, mobility and singularity of parallel mechanism, Mechanism
  and Machine Theory 116 (2017) 273--293.

\bibitem{Chai2017}
X.~Chai, Q.~Li, W.~Ye, {Mobility analysis of overconstrained parallel mechanism
  using Grassmann-Cayley algebra}, Applied Mathematical Modelling 51 (2017)
  643--654.

\bibitem{Yang2020}
S.~Yang, Y.~Li, Classification and analysis of constraint singularities for
  parallel mechanisms using differential manifolds, Applied Mathematical
  Modelling 77 (2020) 469--477.

\bibitem{KimJeongPark15a}
J.~Kim, J.~Jeong, J.~Park, {Inverse kinematics and geometric singularity
  analysis of a 3-SPS/S redundant motion mechanism using conformal geometric
  algebra}, Mechanism and Machine Theory 90 (2015) 23--36.

\bibitem{Huo2008}
L.~Huo, L.~Baron, {The joint-limits and singularity avoidance in robotic
  welding}, Industrial Robot, 35~(5) (2008) 456--464.

\bibitem{Yahya2012}
S.~Yahya, M.~Moghavvemi, H.~Mohamed, Singularity avoidance of a six degree of
  freedom three dimensional redundant planar manipulator, Computers \&
  Mathematics with Applications 64~(5) (2012) 856--868.

\bibitem{Siciliano08a}
B.~Siciliano, L.~Sciavicco, L.~Villani, G.~Oriolo, Robotics: Modelling,
  Planning and Control, Springer Publishing Company, 2008.

\bibitem{Yao2018}
H.~Yao, Q.~Li, Q.~Chen, X.~Chai, {Measuring the closeness to singularities of a
  planar parallel manipulator using geometric algebra}, Applied Mathematical
  Modelling 57 (2018) 192--205.

\bibitem{Nawratil2019}
G.~Nawratil, {Singularity distance for parallel manipulators of Stewart Gough
  type}, in: T.~Uhl (Ed.), Advances in Mechanism and Machine Science, Springer
  International Publishing, 2019, pp. 259--268.

\bibitem{Bu_2016}
W.~Bu, Closeness to singularities of robotic manipulators measured by
  characteristic angles, Robotica 34~(9) (2016) 2105--2115.

\bibitem{DoranLasenby03}
C.~Doran, A.~Lasenby, Geometric Algebra for Physicists, Cambridge University
  Press, 2003.

\bibitem{Dorst-Fontijne-Mann-2007}
L.~Dorst, D.~Fontijne, S.~Mann, {Geometric algebra for computer science: An
  object-oriented approach to geometry}, Morgan Kaufmann Publishers Inc., 2007.

\bibitem{Clifford82}
W.~Clifford, H.~Smith, R.~Tucker, Mathematical Papers by William Kingdon
  Clifford -- Edited, Macmillan London, 1882.

\bibitem{MurrayLiShankarSastry94}
R.~Murray, Z.~Li, S.~Shankar-Sastry, A Mathematical Introduction to Robotic
  Manipulation, CRC Press, 1994.

\bibitem{DavidsonHunt04}
J.~Davidson, K.~Hunt, Robots and Screw Theory: Applications of Kinematics and
  Statics to Robotics, Oxford University Press, 2004.

\bibitem{Tsai99}
L.~Tsai, Robot Analysis: The Mechanics of Serial and Parallel Manipulators,
  John Wiley and Sons, 1999.

\bibitem{Sangwine17}
S.~Sangwine, E.~Hitzer, {Clifford Multivector Toolbox (for MATLAB)}, Advances
  in Applied Clifford Algebras 27~(1) (2017) 539--558.

\bibitem{ZaplanaClaretBasanez17}
I.~Zaplana, J.~Claret, L.~Basanez, {Kinematic analysis of redundant robotic
  manipulators: applications to Kuka LWR 4+ and ABB Yumi}, Revista
  Iberoamericana de Autom\'{a}tica e Inform\'{a}tica Industrial 15~(2) (2018)
  192--202.

\bibitem{LavorXamboZaplana18}
C.~Lavor, S.~Xamb\'{o}-Descamps, I.~Zaplana, {A Geometric Algebra Invitation to
  Space-Time Physics, Robotics and Molecular Geometry}, SRMA/Springerbriefs,
  Springer, 2018.

\bibitem{Basanez09}
L.~Basa{\~{n}}ez, R.~Su{\'a}rez, Teleoperation, in: S.~Nof (Ed.), Springer
  Handbook of Automation, Springer Berlin Heidelberg, 2009, pp. 449--468.

\end{thebibliography}

%\clearpage

%\section*{Author Biography}

%\begin{biography}{\includegraphics[width=66pt,height=86pt,draft]{empty}}{\textbf{Author Name.} This is sample author biography text this is sample author biography text this is sample author biography text this is sample author biography text this is sample author biography text this is sample author biography text this is sample author biography text this is sample author biography text this is sample author biography text this is sample author biography text this is sample author biography text this is sample author biography text this is sample author biography text this is sample author biography text this is sample author biography text this is sample author biography text this is sample author biography text this is sample author biography text this is sample author biography text this is sample author biography text this is sample author biography text.}
%\end{biography}

\end{document}